%% file: main.tex
\documentclass{article}

\PassOptionsToPackage{numbers, compress}{natbib}
\usepackage[preprint]{ewrl_2024}

\usepackage[utf8]{inputenc} % allow utf-8 input
\usepackage[T1]{fontenc}    % use 8-bit T1 fonts
\usepackage{hyperref}       % hyperlinks
\usepackage{url}            % simple URL typesetting
\usepackage{booktabs}       % professional-quality tables
\usepackage{amsfonts}       % blackboard math symbols
\usepackage{nicefrac}       % compact symbols for 1/2, etc.
\usepackage{microtype}      % microtypography
\usepackage{xcolor}         % colors
\usepackage{amssymb}        % Defines common symbols like \mathbb R
\usepackage{mathtools}      % Extends amsmath, providing common math tools
\usepackage{mathrsfs}       % Enables \mathscr, which can work in cases that \mathcal does not
\mathtoolsset{showonlyrefs} % Only number equations that are referenced (optional)
\usepackage{graphicx}       % For including images
\usepackage{subcaption}     % Allows for the use of subfigures and subcaptions
\usepackage[space]{grffile} % For spaces in image names
\usepackage{url}            % For displaying urls

\usepackage{float}
\usepackage{algorithm}
\usepackage{algorithmic}

\title{AFU: Actor-Free critic Updates in \mbox{off-policy} RL for \mbox{continuous} \mbox{control}}

\author{Nicolas Perrin-Gilbert  \\
    Sorbonne Université, CNRS, Institut des Systèmes Intelligents et de Robotique, ISIR,\\ 
    F-75005 Paris, France \\
    \texttt{nicolas.perrin-gilbert@cnrs.fr} \\
}

\begin{document}

\maketitle

\begin{abstract}
This paper presents AFU\footnote{An open-source JAX-based implementation of AFU is available at: \url{www.github.com/perrin-isir/afu}.}, an off-policy deep RL algorithm addressing in a new way the challenging ``max-Q problem'' in Q-learning for continuous action spaces, with a solution based on regression and conditional gradient scaling. AFU has an actor but its critic updates are entirely independent from it. As a consequence, the actor can be chosen freely. In the initial version, AFU-alpha, we employ the same stochastic actor as in Soft Actor-Critic (SAC), but we then study a simple failure mode of SAC and show how AFU can be modified to make actor updates less likely to become trapped in local optima, resulting in a second version of the algorithm, AFU-beta. Experimental results demonstrate the sample efficiency of both versions of AFU, marking it as the first model-free off-policy algorithm competitive with state-of-the-art actor-critic methods while departing from the actor-critic perspective.
\end{abstract}

%%%%%%%%%%%%%%%%%%%%%%%%%%%%%%%%%%%%%%%%%%%%%%%%%%%%%%%%%%%%%%%%
%% Section: Introduction
%%%%%%%%%%%%%%%%%%%%%%%%%%%%%%%%%%%%%%%%%%%%%%%%%%%%%%%%%%%%%%%%
\section{Introduction}\label{sec:intro}
\input{intro}

%%%%%%%%%%%%%%%%%%%%%%%%%%%%%%%%%%%%%%%%%%%%%%%%%%%%%%%%%%%%%%%%
%% Section: Related Work
%%%%%%%%%%%%%%%%%%%%%%%%%%%%%%%%%%%%%%%%%%%%%%%%%%%%%%%%%%%%%%%%
\section{Related Work}\label{sec:rw}
\input{rw}

%%%%%%%%%%%%%%%%%%%%%%%%%%%%%%%%%%%%%%%%%%%%%%%%%%%%%%%%%%%%%%%%
%% Section: Preliminaries
%%%%%%%%%%%%%%%%%%%%%%%%%%%%%%%%%%%%%%%%%%%%%%%%%%%%%%%%%%%%%%%%
\section{Preliminaries}\label{sec:prelim}
\input{prelim}

%%%%%%%%%%%%%%%%%%%%%%%%%%%%%%%%%%%%%%%%%%%%%%%%%%%%%%%%%%%%%%%%
%% Section: Solving the max-Q problem without regularization
%%%%%%%%%%%%%%%%%%%%%%%%%%%%%%%%%%%%%%%%%%%%%%%%%%%%%%%%%%%%%%%%
\section{A new way to solve the max-Q problem}\label{sec:maxq}
\input{maxq}

%%%%%%%%%%%%%%%%%%%%%%%%%%%%%%%%%%%%%%%%%%%%%%%%%%%%%%%%%%%%%%%%
%% Section: Training the Actor
%%%%%%%%%%%%%%%%%%%%%%%%%%%%%%%%%%%%%%%%%%%%%%%%%%%%%%%%%%%%%%%%
\section{Actor-free critic updates and actor training}\label{sec:actor}
\input{actor}

%%%%%%%%%%%%%%%%%%%%%%%%%%%%%%%%%%%%%%%%%%%%%%%%%%%%%%%%%%%%%%%%
%% Section: AFU-alpha
%%%%%%%%%%%%%%%%%%%%%%%%%%%%%%%%%%%%%%%%%%%%%%%%%%%%%%%%%%%%%%%%
\section{AFU-alpha}\label{sec:alpha}
\input{afualpha}

%%%%%%%%%%%%%%%%%%%%%%%%%%%%%%%%%%%%%%%%%%%%%%%%%%%%%%%%%%%%%%%%
%% Section: Training the Actor
%%%%%%%%%%%%%%%%%%%%%%%%%%%%%%%%%%%%%%%%%%%%%%%%%%%%%%%%%%%%%%%%
\section{A simple failure mode of SAC}\label{sec:failure}
\input{failureSAC}

%%%%%%%%%%%%%%%%%%%%%%%%%%%%%%%%%%%%%%%%%%%%%%%%%%%%%%%%%%%%%%%%
%% Section: Training the Actor
%%%%%%%%%%%%%%%%%%%%%%%%%%%%%%%%%%%%%%%%%%%%%%%%%%%%%%%%%%%%%%%%
\section{AFU-beta}\label{sec:beta}
\input{afubeta}

%%%%%%%%%%%%%%%%%%%%%%%%%%%%%%%%%%%%%%%%%%%%%%%%%%%%%%%%%%%%%%%%
%% Section: Conclusion
%%%%%%%%%%%%%%%%%%%%%%%%%%%%%%%%%%%%%%%%%%%%%%%%%%%%%%%%%%%%%%%%
\section{Conclusion}\label{sec:ccl}
\input{ccl}

\section*{Acknowledgements}\label{sec:ack}
This work was supported by the French National Research Agency (ANR), Project ANR-18-CE33-0005 HUSKI,
and has received funding from the European Commission's Horizon Europe Framework Program under
grant agreement No 101070381 (PILLAR-Robots project). For the numerical experiments, this work 
was granted access to the HPC resources of IDRIS under the 
allocations 2022-A0131013011 and 2023-A0151013011 made by GENCI.

%\subsection{Tables}
%\label{sec:tables}
%Tables must be centred, neat, clean and legible. Do not use hand-drawn tables. The table number and title always appear after the table. See Table~\ref{tab:exampleTable}. Place one line space before the table title, one line space after the table title, and one line space after the table. Tables are numbered consecutively.
%
%\begin{table}[htbp]
%    \begin{center}
%        \begin{tabular}{ll}
%            \multicolumn{1}{l}{\bf PART}  &\multicolumn{1}{l}{\bf DESCRIPTION}
%            \\ \hline \\
%            Actor         &Stores and updates the policy \\
%            Critic        &Stores and updates a value function \\
%        \end{tabular}
%    \end{center}
%    \caption{Sample table caption}
%    \label{tab:exampleTable}
%\end{table}

%%%%%%%%%%%%%%%%%%%%%%%%%%%%%%%%%%%%%%%%%%%%%%%%%%%%%%%%%%%%%%%%
%% NOTE: THIS MARKS THE END OF THE "MAIN TEXT"
%%%%%%%%%%%%%%%%%%%%%%%%%%%%%%%%%%%%%%%%%%%%%%%%%%%%%%%%%%%%%%%%

%%%%%%%%%%%%%%%%%%%%%%%%%%%%%%%%%%%%%%%%%%%%%%%%%%%%%%%%%%%%%%%%
%% Bibliography
%%%%%%%%%%%%%%%%%%%%%%%%%%%%%%%%%%%%%%%%%%%%%%%%%%%%%%%%%%%%%%%%
\newpage

\bibliographystyle{plain}
\bibliography{main}

%%%%%%%%%%%%%%%%%%%%%%%%%%%%%%%%%%%%%%%%%%%%%%%%%%%%%%%%%%%%%%%%
%% Appendices
%%%%%%%%%%%%%%%%%%%%%%%%%%%%%%%%%%%%%%%%%%%%%%%%%%%%%%%%%%%%%%%%
\newpage
\appendix
%
%\section{The first appendix}
%\label{sec:appendix1}
%This is an example of an appendix. 
\section{Hyperparameters}
\label{app:hypers}
The following hyperparameters were used in all our experiments.

We did not do any reward scaling.

\begin{table}[htbp]
    \begin{center}
        \begin{tabular}{ll}
            \multicolumn{2}{c}{\bf AFU, SAC \& TD3 Hyperparameters}
            \\ \hline \\
            optimizer &Adam \citep{kingma2014adam}\\
            actor learning rate         & $3 \cdot 10^{-4}$ \\
            critic learning rate        & $3 \cdot 10^{-4}$ \\
            temperature learning rate \textbf{(only AFU \& SAC)}        & $3 \cdot 10^{-4}$ \\
            discount ($\gamma$) &  0.99\\
            replay buffer size & $10^6$\\
            initial steps with random actions & $10^4$\\
            number of hidden layers (all networks) & 2\\
            number of hidden units per layer & 256\\
            number of samples per mini-batch & 256\\
            nonlinearity & ReLU\\
            target smoothing coefficient ($\tau$) & 0.01\\
            target update interval & 1\\
            policy update interval \textbf{(only TD3)} & 2\\
            exploration noise standard deviation \textbf{(only TD3)} & 0.2\\
            noise clipping \textbf{(only TD3)} & 0.5\\            
            target entropy \textbf{(only AFU \& SAC)} & $-d$ ($d=$ action space dimension)\\
            initial temperature \textbf{(only AFU \& SAC)} & $1$ \\
            max. actor log std (before tanh) \textbf{(only AFU \& SAC)} & $2$\\
            min. actor log std (before tanh) \textbf{(only AFU \& SAC)} & $-10$\\
        \end{tabular}
    \end{center}
\end{table}

\section{Conditional gradient rescaling seen as adaptive regularization}
\label{app:rescaling}

Let us denote by $e(s,a)$ the error: 
$$
e(s,a) = V_{\varphi_i}(s) + A_{\xi_i}(s,a) - Q_\psi(s, a),
$$
and let us assume that this error is negative (otherwise our method simply applies a standard gradient descent step).
We denote by $e'(s,a)$ the following term: 
$$
e'(s,a) = (1 - \varrho) V_{\varphi_i}(s) + \varrho V_{\varphi^{\text{no\_grad}}_i}(s) + A_{\xi_i}(s,a) - Q_\psi(s, a),
$$
which is equal to $e(s,a)$ in value but has different gradients.

Our method applies a gradient descent step on $e'(s,a)^2$ for both $\varphi_i$ and $\xi_i$. 
For $\xi_i$, the gradient is the same as for $e(s,a)^2$ and $e'(s,a)^2$.
For $\varphi_i$, the gradients are: 
$$
\nabla_{\varphi_i}e(s,a)^2 = 2e(s,a) \nabla_{\varphi_i} V_{\varphi_i}(s),
$$
$$
\nabla_{\varphi_i}e'(s,a)^2 = (1 - \varrho) 2e(s,a) \nabla_{\varphi_i} V_{\varphi_i}(s).
$$
Let us define $e^{\text{no\_grad}}(s,a)$: a ``frozen'' version of $e(s,a)$ leading to no gradients at all (i.e. relying on copies of both $\varphi_i$ and $\xi_i$).
Our method is equivalent to the application of a gradient step (for both $\varphi_i$ and $\xi_i$) to the following term: 
$$
e(s,a)^2 - 2\varrho e^{\text{no\_grad}} V_{\varphi_i}(s) = e(s,a)^2 + 2\varrho |e^{\text{no\_grad}}(s,a)| V_{\varphi_i}(s),
$$
where we see the squared error and a simple regularization term that penalizes large values of $V_{\varphi_i}(s)$ (thus putting a ``downward pressure'' on 
$V_{\varphi_i}(s)$). As a result, the proposed conditional gradient rescaling method can be understood as an adaptive regularization scheme 
in which the regularization weight is proportional to the absolute value of the error. 
It means that the convergence of $V_{\varphi_i}(s)$ toward $\max_{a \in A} (Q_\psi(s, a))$ is not theoretically guaranteed: if the regression quickly converges to an exact solution, the gradients vanish and $V_{\varphi_i}(s)$ can remain strictly greater than the true maximum. However, if a non negligible error remains, the adaptive regularization is effective and $V_{\varphi_i}(s)$ progressively decreases toward an approximation of the true maximum, which is what we observe in practice.

\section{Constraining the sign of $A_{\xi_i}(s,a)$ in a soft way}
\label{app:soft}

We let $A_{\xi_i}$ possibly return positive outputs, but we modify the regression loss to have only non-positive \emph{targets} for $A_{\xi_i}(s,a)$. In Equation~\eqref{eq:LambdaVA}, we call $Q_\psi(s, a) - \Upsilon_{i}^{a}(s)$ the \emph{target} of $A_{\xi_i}(s,a)$, as it is the value of $A_{\xi_i}(s,a)$ minimizing $\Bigl(\Upsilon_{i}^{a}(s) + A_{\xi_i}(s,a) - Q_\psi(s, a) \Bigr)^2$. If \mbox{$Q_\psi(s, a) - \Upsilon_{i}^{a}(s) > 0$}, the best non-positive \emph{target} for $A_{\xi_i}(s,a)$ is $0$, in which case the \emph{target} for \mbox{$\Upsilon_{i}^{a}(s) - Q_\psi(s, a)$} should also be $0$. In this situation, we replace \mbox{$\Bigl(\Upsilon_{i}^{a}(s) + A_{\xi_i}(s,a) - Q_\psi(s, a) \Bigr)^2$} by \mbox{$\Bigl(\Upsilon_{i}^{a}(s) - Q_\psi(s, a) \Bigr)^2 + \Bigl(A_{\xi_i}(s,a)\Bigr)^2$}. To do so, we introduce $Z$:
\begin{align}
Z(x,y) =\left\{
    \begin{array}{ll}
      (x+y)^2, & \mbox{if $x \geq 0$}.\\
      x^2+y^2, & \mbox{otherwise}.
    \end{array}
  \right.
\end{align}
The loss of Equation~\ref{eq:LambdaVA} is updated as follows:
\begin{align}
\Lambda'_{V, A}(\varphi_i, \xi_i) = \underset{(s, a, \_, \_) \in B}{\text{Mean}}\left[  Z\Bigl(\Upsilon_{i}^{a}(s) - Q_\psi(s, a), A_{\xi_i}(s,a) \Bigr) \right].
\end{align}

\section{Experiments on a toy max-Q problem}
\label{app:toy}

We empirically compare our method to 3 baselines (IQL, SQL and EQL) on a toy problem.
%but we can get some intuition by empirically studying a much simpler case where $V_{\varphi_i}(s)$ is replaced by a single variable $x$, $Q_\psi(s, a)$ by two values $0$ and $-1$ (as if there were only a single state and two actions), and $A_{\xi_i}(s,a)$ by two values governed by a single variable: $y$ and $y-1-\epsilon$, with a fixed $\epsilon > 0$. The objective is to match $0$ with $x + y$ and $-1$ with $x+y-1-\epsilon$. We consider mini-batches of size 1, so at every step, we randomly select either the loss associated to $x+y$ and the target $0$, or the loss associated to $x+y-1-\epsilon$ and the target $-1$.  One gradient descent update is made for both $x$ and $y$ at every step. Since $\epsilon \neq 0$, there is no perfect solution, so the gradient descent produces oscillations that enable the asymmetry introduced by $\varrho$ to cause the convergence of $x$. With a learning rate slowly decreasing toward $0$, we observe that $x$ converges toward $0$, i.e. the maximum of $0$ and $-1$, and $y$ converges toward $\epsilon/2$. The convergence rate is faster for large values of $\epsilon$, which indicates that not converging early to perfect solutions is essential for the efficiency of the proposed method.
We define the function $Q_{toy}(s,a) = \sin(4s) + 0.7\cos(4a)$
for $s \in [-1, 1]$ and $a \in [-1, 1]$.
We use a single feedforward neural network for $V$ ($V_\varphi$) and a single feedforward neural network for $A$ ($A_\xi$). Both networks have two hidden layers of size 256 and ReLU activations in the hidden layers. Our method trains both $V_\varphi$ and $A_\xi$, while the 3 baselines IQL, SQL and EQL directly train $V_\varphi$. All 3 baselines have been successfully applied to offline reinforcement learning.
 
SQL and EQL are derived in \cite{DBLP:conf/iclr/Xu0LYWCZ23} from a general method called Implicit Value Regularization. It relies on a behavior-regularized MDP with a term that penalizes policies diverging from the underlying behavior policy of the training dataset. Various f-divergences can be used to measure the difference between the policy and the behavior policy, resulting in distinct algorithms, including SQL and EQL which are special cases. They have distinct losses for the training of $V_\varphi(s)$, both depending on a parameter $\alpha$, and in both cases, for $\alpha \rightarrow 0$, $V_\varphi(s)$ is trained to approximate the maximum operator over in-support values, i.e. $\max_{a \in A} (Q_{toy}(s, a))$.
However, similarly to IQL, very small values of $\alpha$ result in unbalanced losses, so in practice the values leading to the best results on the benchmarks tested in \cite{DBLP:conf/iclr/Xu0LYWCZ23} are $\alpha=0.1$, $\alpha=0.5$, $\alpha=1$ and $\alpha=3$ for SQL, and $\alpha=0.5$, $\alpha=2.0$ and $\alpha=5$ for EQL. 

For IQL, SQL and EQL, since unbalanced losses are not an issue on this simple toy problem, we include parameters leading to a better resolution of the max-Q problem, but that are not representative of the parameters working well in actual offline RL experiments. We observe that, with our method, although $\varrho=0.05$ leads to overestimations, for a wide range of parameter values (from $\varrho=0.2$ to $\varrho=0.7$), we obtain more accurate results than with all the other baselines, even when considering parameter values that are inapplicable to offline RL. Besides, with our proposed approach, the different values of $\varrho$ that perform well do not result in unbalanced losses.

\begin{figure}[H]
    \begin{subfigure}{0.5\linewidth}
        \includegraphics[width=\linewidth]{images/our_method}
        \caption{$V_{\varphi}(s)$ is trained jointly with $A_{\xi}(s,a)$ by iterating gradient descent steps on the loss $\Lambda'_{V, A}(\varphi, \xi)$ described by Equation~\eqref{eq:LambdaVAmodif}. $\varrho \in [0.2, 0.7]$ results in precise approximations of $s \mapsto \max_{a \in A} (Q_\psi(s, a))$.}
        \label{fig:toyproblem_sub1}
    \end{subfigure}
    \hspace*{\fill}
    \begin{subfigure}{0.5\linewidth}
        \includegraphics[width=\linewidth]{images/iql}
        \caption{Results of the training with the loss from IQL \citep{DBLP:conf/iclr/KostrikovNL22} for 4 different values of the hyperparameter $\tau$. Values used in actual (offline) RL experiments are not greater than $0.9$.}
        \label{fig:toyproblem_sub2}
    \end{subfigure}
    \vskip\baselineskip
        \begin{subfigure}{0.5\linewidth}
        \includegraphics[width=\linewidth]{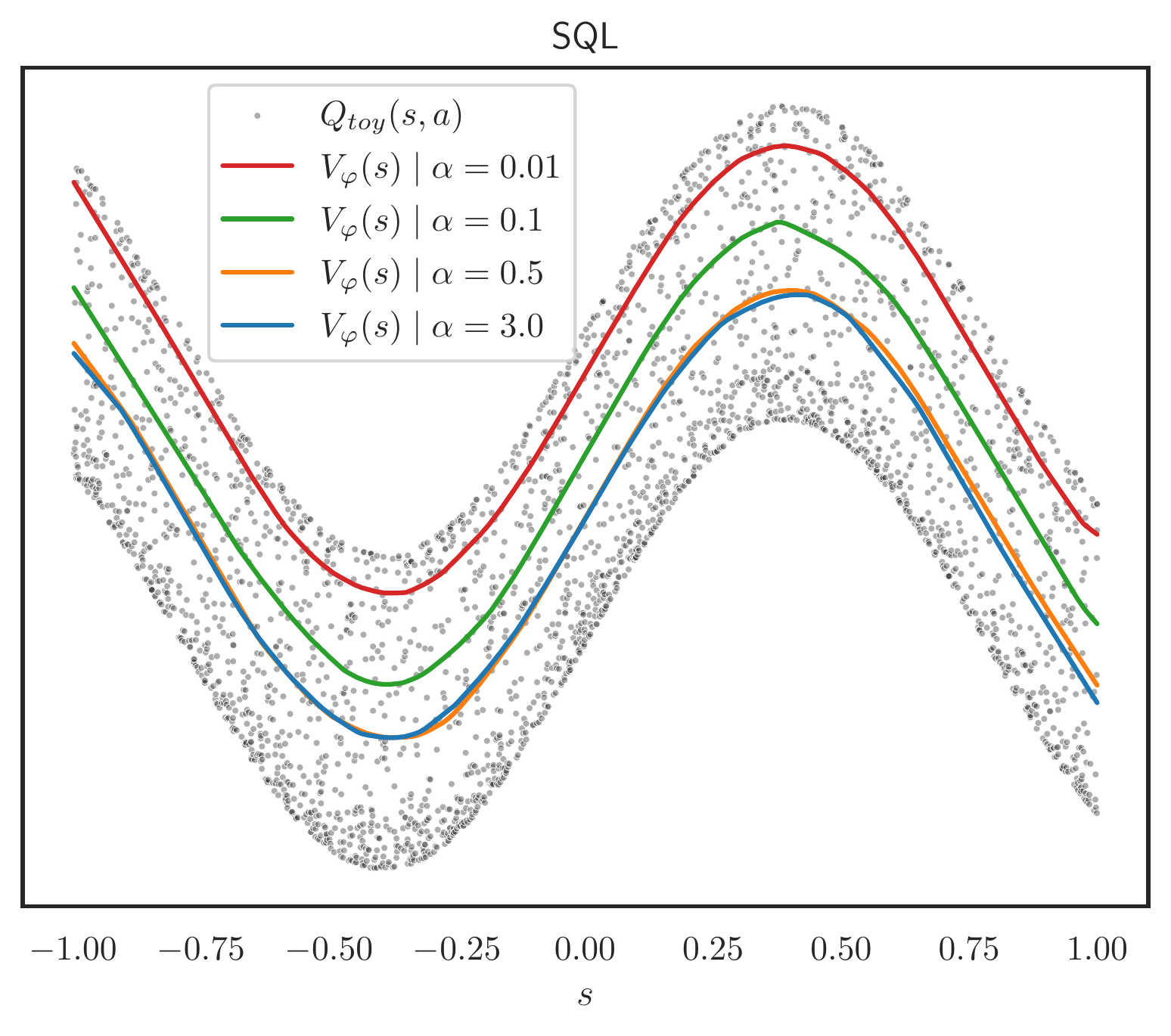}
        \caption{Results of the training with the loss from SQL \citep{DBLP:conf/iclr/Xu0LYWCZ23} for 4 different values of the hyperparameter $\alpha$. Values used in actual (offline) RL experiments are not smaller than $0.1$.}
        \label{fig:toyproblem_sub3}
    \end{subfigure}
    \hspace*{\fill}
    \begin{subfigure}{0.5\linewidth}
        \includegraphics[width=\linewidth]{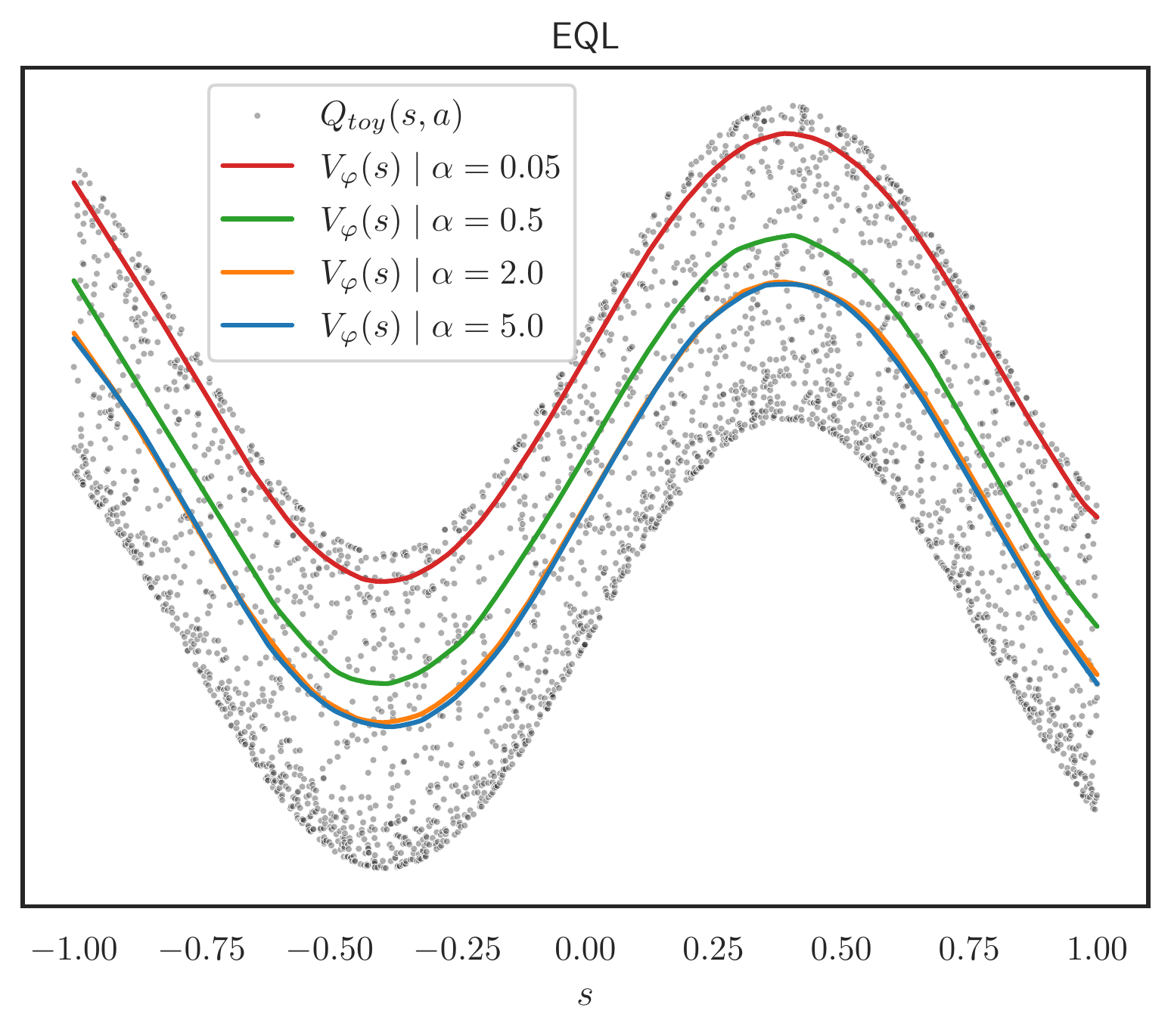}
        \caption{Results of the training with the loss from EQL \citep{DBLP:conf/iclr/Xu0LYWCZ23} for 4 different values of the hyperparameter $\alpha$. Values used in actual (offline) RL experiments are not smaller than $0.5$.}
        \label{fig:toyproblem_sub4}
    \end{subfigure}
    \caption{$Q_{toy}(s,a) = \sin(4s) + 0.7\cos(4a)$
for $s \in [-1, 1]$ and $a \in [-1, 1]$. We compare our method to IQL, SQL and EQL which all train $V_{\varphi}(s)$ to approximate the function $s \mapsto \max_{a \in A} (Q_{toy}(s, a))$, i.e. solve the max-Q problem. All trainings are done with 3000 gradient descent steps. At each step, a loss is computed on a batch composed of 256 uniformly randomly drawn values of $s$ and $a$.}
    \label{fig:toyproblem_appendix}
\end{figure}

\section{SAC-like actor training}
\label{app:sac}

Let $\pi_\theta$ denote the actor. We follow a common implementation in which its backbone is a feedforward neural network returning action distributions as state-dependent Gaussians with diagonal covariance matrices. Since actions are usually constrained between $-1$ and $1$, we apply a $\tanh$ transformation to its outputs. Given a state $s$, the resulting probability density function is $\pi_\theta(\cdot | s)$. The actor $\pi_\theta$ can transform input noise vectors sampled from a fixed distribution into action samples. Again, we train $\pi_\theta$ on mini-batches of transitions. We use the actor to resample an action $a_s$ for each state $s$ of a mini-batch $B$. The actor loss $L_{\pi}(\theta)$ is based on the average Kullback-Leibler divergence between the actor's output distributions and targeted Boltzmann policy distributions. It is defined as follows:
\begin{align}
L_{\pi}(\theta) = \underset{\substack{(s, \_, \_, \_) \in B \\ a_s \sim \pi_\theta(\cdot | s)}}{\text{Mean}}\Big[ \alpha \log(\pi_\theta(a_s | s)) - Q_\psi(s, a_s) \Big],
\end{align}
where $\alpha$ is a temperature parameter. As in SAC, we adjust this temperature via gradient descent on a loss aiming at keeping the average entropy of action distributions close to a target entropy $\bar{\mathcal{H}}$:
\begin{align}
L_{\text{temp}}(\alpha) = \underset{\substack{(s, \_, \_, \_) \in B \\ a_s \sim \pi_\theta(\cdot | s)}}{\text{Mean}}\Big[ -\alpha \log(\pi_\theta(a_s | s)) - \alpha \bar{\mathcal{H}} \Big].
\end{align}

\section{Experimental results for AFU-beta}
\label{app:beta}

\begin{figure}[H]
    \begin{subfigure}{0.32\linewidth}
        \includegraphics[width=\linewidth]{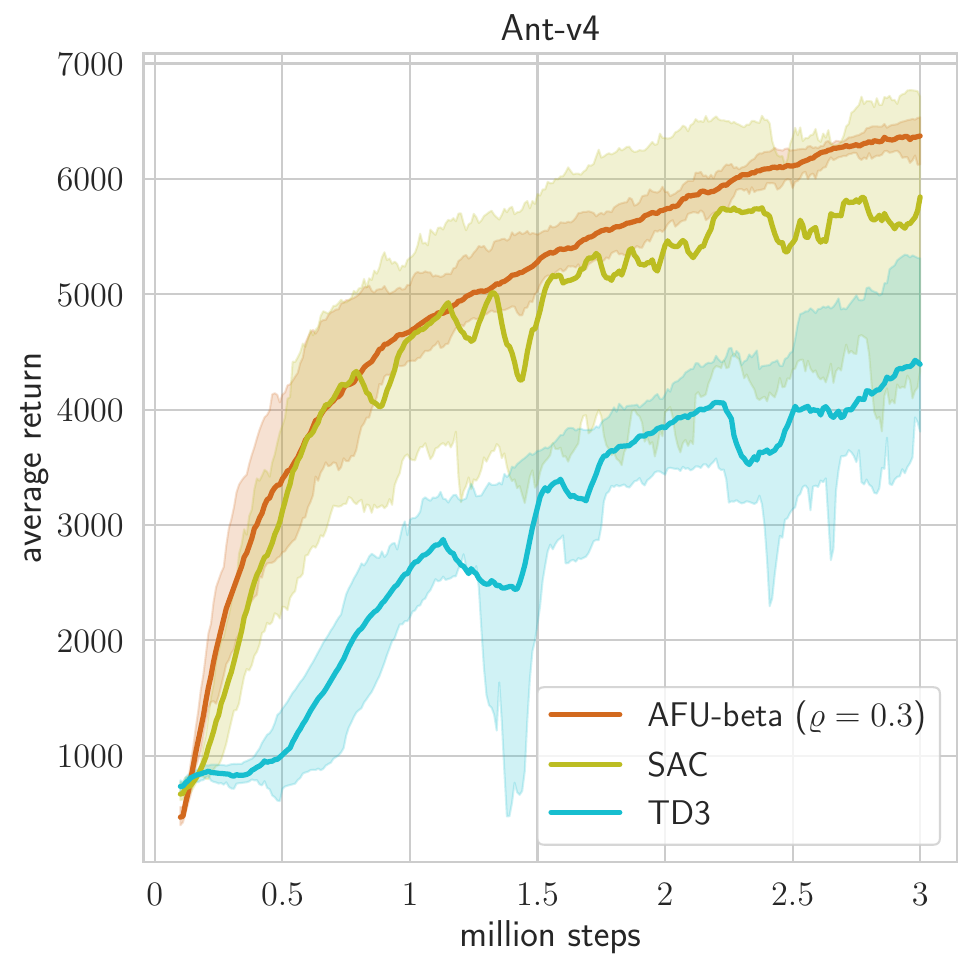}
        \label{fig:Ant}
    \end{subfigure}
    \hspace*{\fill}
    \begin{subfigure}{0.32\linewidth}
        \includegraphics[width=\linewidth]{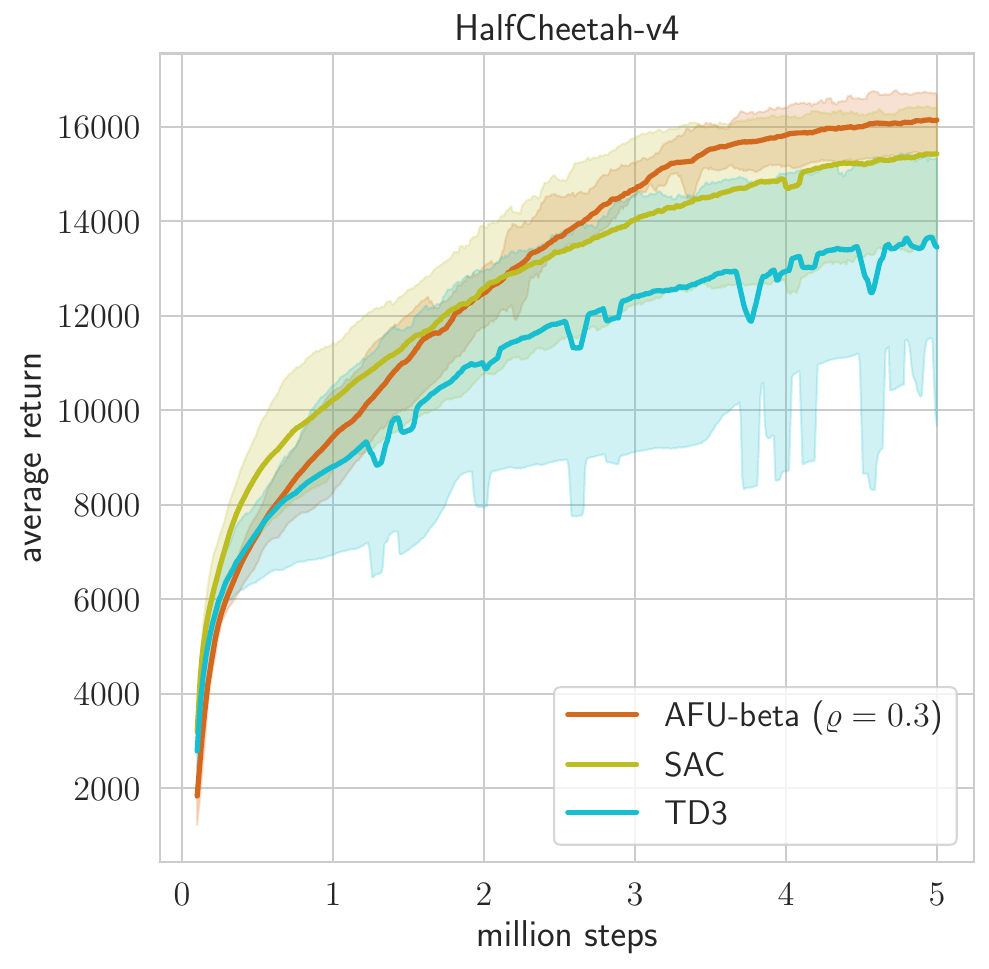}
        \label{fig:HC}
    \end{subfigure}
     \hspace*{\fill}
    \begin{subfigure}{0.32\linewidth}
        \includegraphics[width=\linewidth]{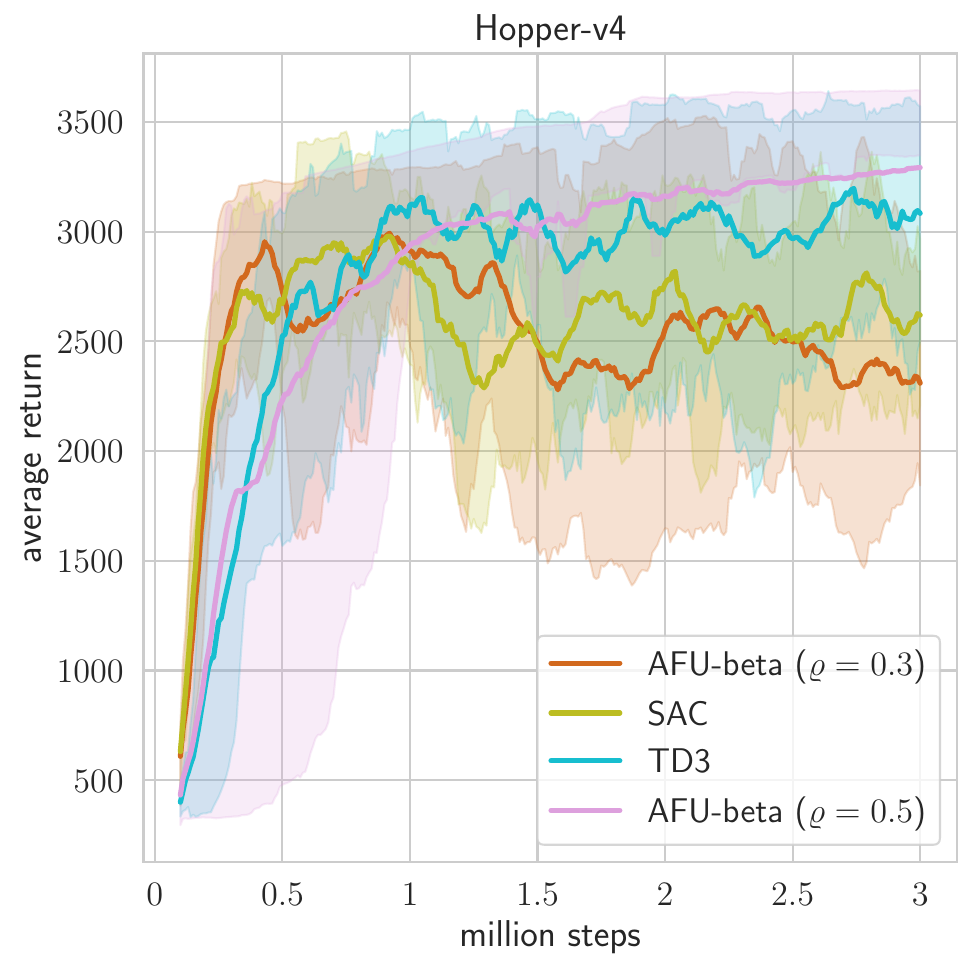}
        \label{fig:Hopper}
    \end{subfigure}
    \vskip\baselineskip
    \begin{subfigure}{0.32\linewidth}
        \includegraphics[width=\linewidth]{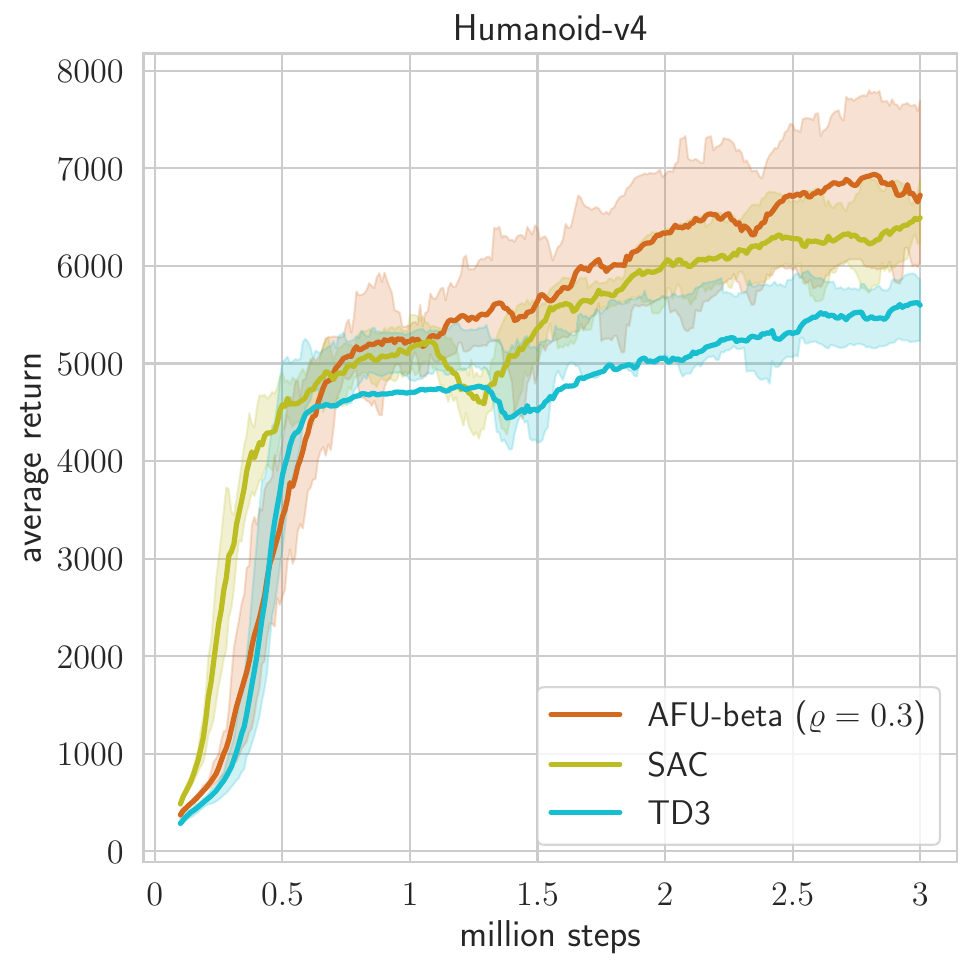}
        \label{fig:Humanoid}
    \end{subfigure}
        \hspace*{\fill}
    \begin{subfigure}{0.32\linewidth}
        \includegraphics[width=\linewidth]{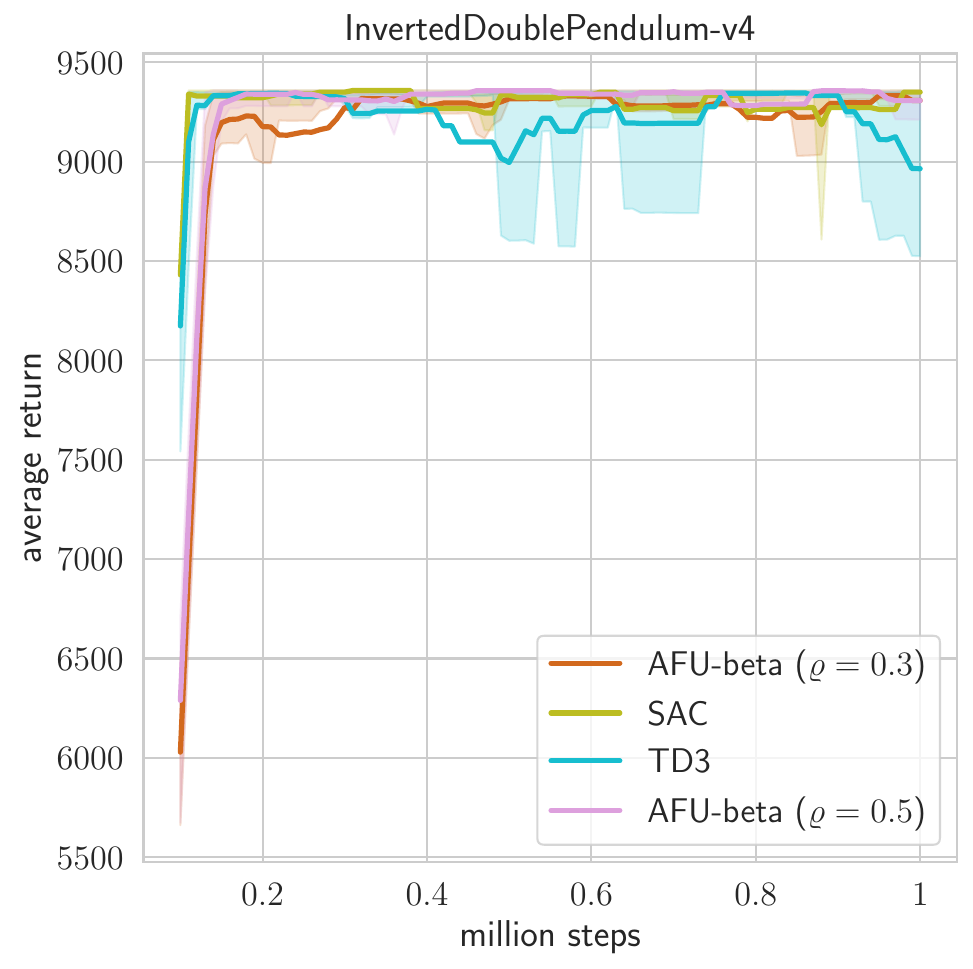}
        \label{fig:IDP}
    \end{subfigure}
    \hspace*{\fill}
    \begin{subfigure}{0.32\linewidth}
        \includegraphics[width=\linewidth]{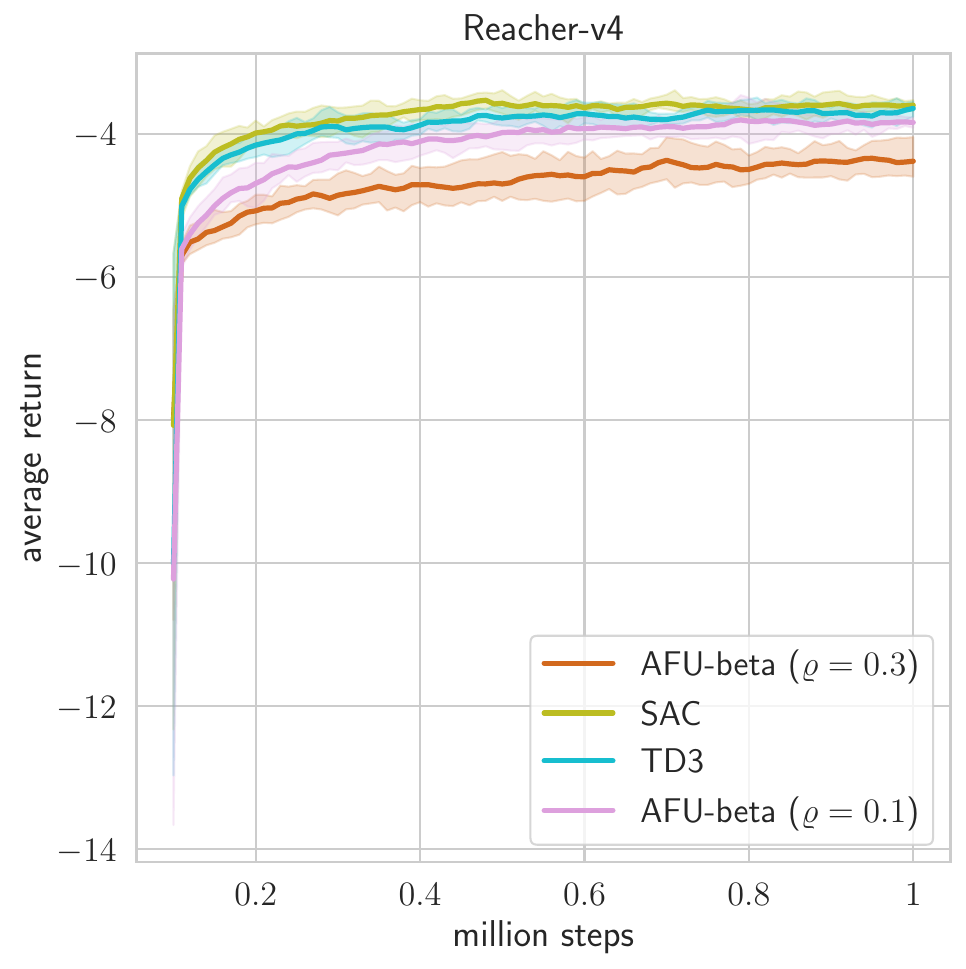}
        \label{fig:Reacher}
    \end{subfigure}
    \vskip\baselineskip
    \begin{subfigure}{0.32\linewidth}
        \includegraphics[width=\linewidth]{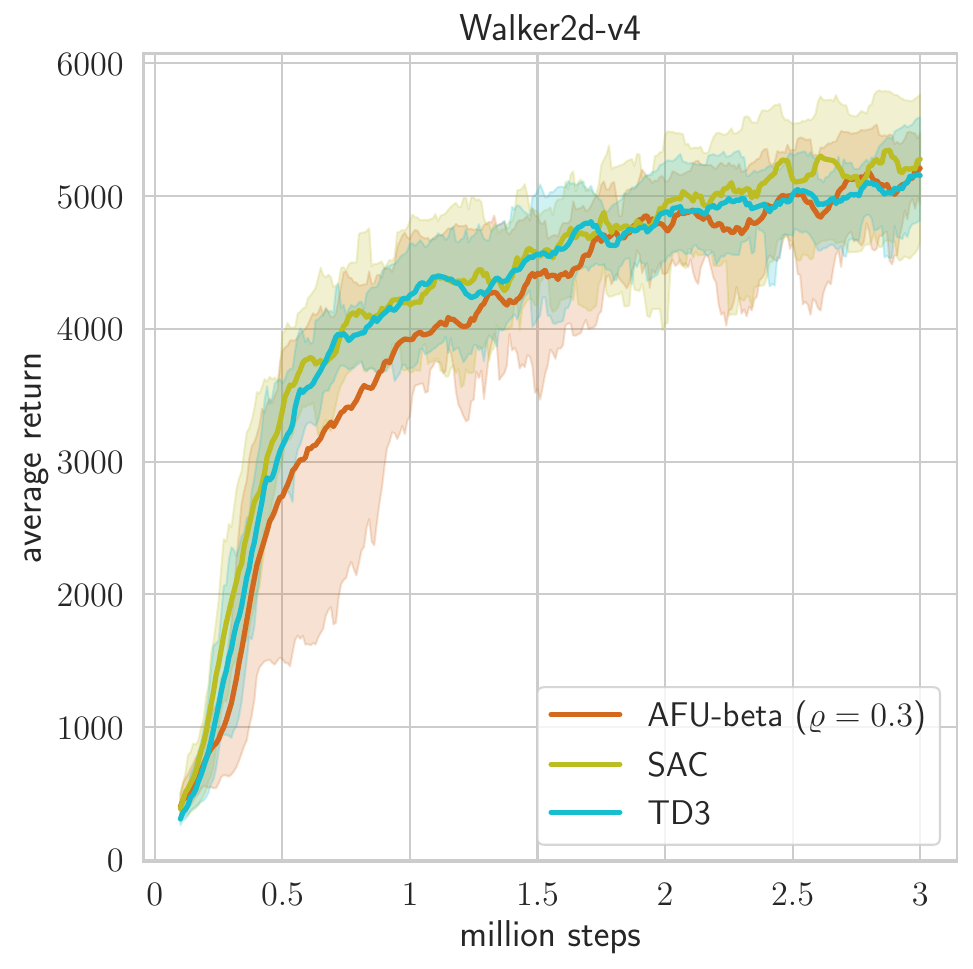}
        \label{fig:Walker}
    \end{subfigure}   
    \hspace*{\fill}
    \begin{subfigure}{0.32\linewidth}
        \includegraphics[width=\linewidth]{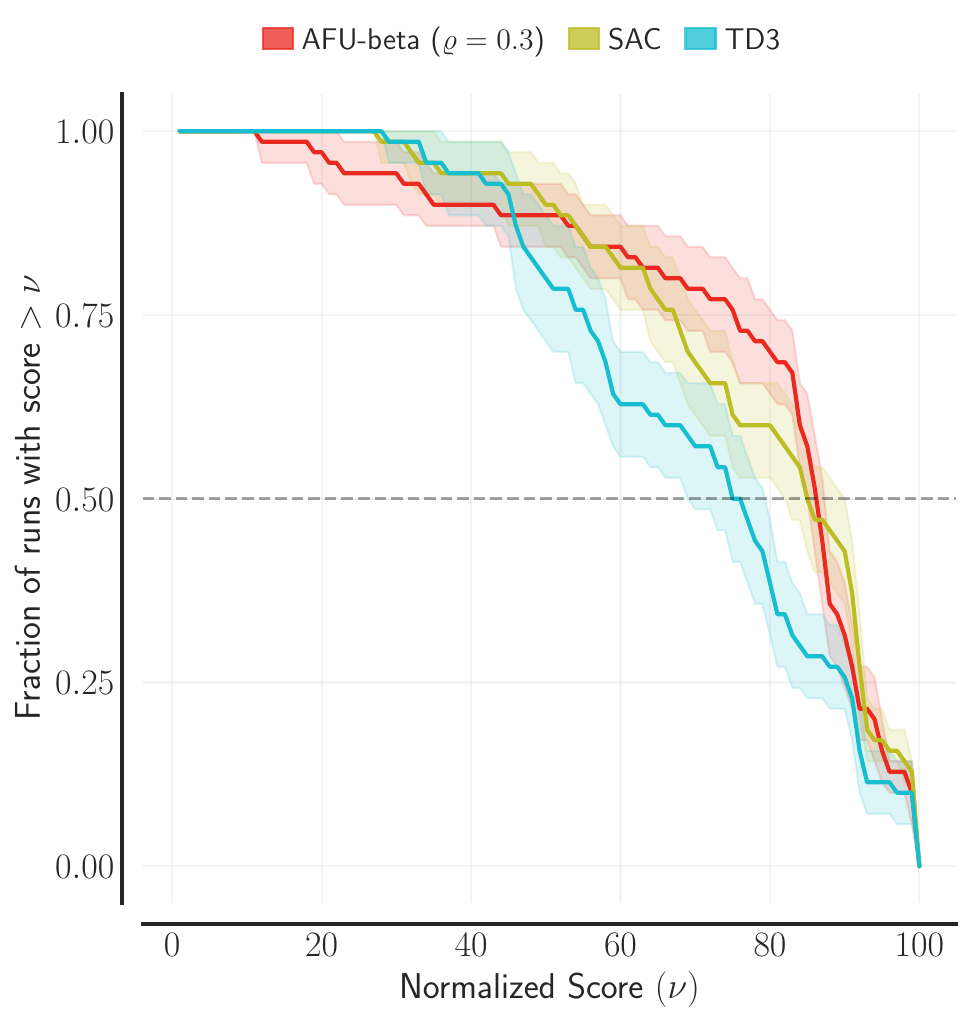}
        \label{fig:perf_profile}
    \end{subfigure}
 
    \caption{Experimental evaluation of AFU-beta for $\varrho = 0.3$ on 7 MuJoCo tasks. We show results with other values of $\varrho$ (among $\{0.1, 0.2, 0.4, 0.5\}$) for tasks in which one of the other values performed significantly better than $0.3$. Results are averaged over 10 runs with different random seeds, and the shaded areas range from the 25th to the 75th percentile. The performance profile plot at the bottom right summarizes results and shows that AFU-beta is competitive with SAC and TD3.}
    \label{fig:Benchmark}
\end{figure}

\section{Learning curves}
\label{app:lc}

The plots below show learning curves for AFU-alpha and AFU-beta for all the values of the hyperparameter $\varrho$ (in $\{0.1, 0.2, 0.3, 0.4, 0.5\}$).

All learning curves are averaged over 10 runs with different random seeds, and the shaded areas range from the 25th to the 75th percentile. For each run, evaluations are done over 10 rollouts every 10,000 steps, and each run is smoothed with a moving average window of size 10. The first 10,000 steps are always done without gradient steps and with uniformly randomly drawn actions.

\begin{figure}[H]
    \begin{subfigure}{0.5\linewidth}
        \includegraphics[width=\linewidth]{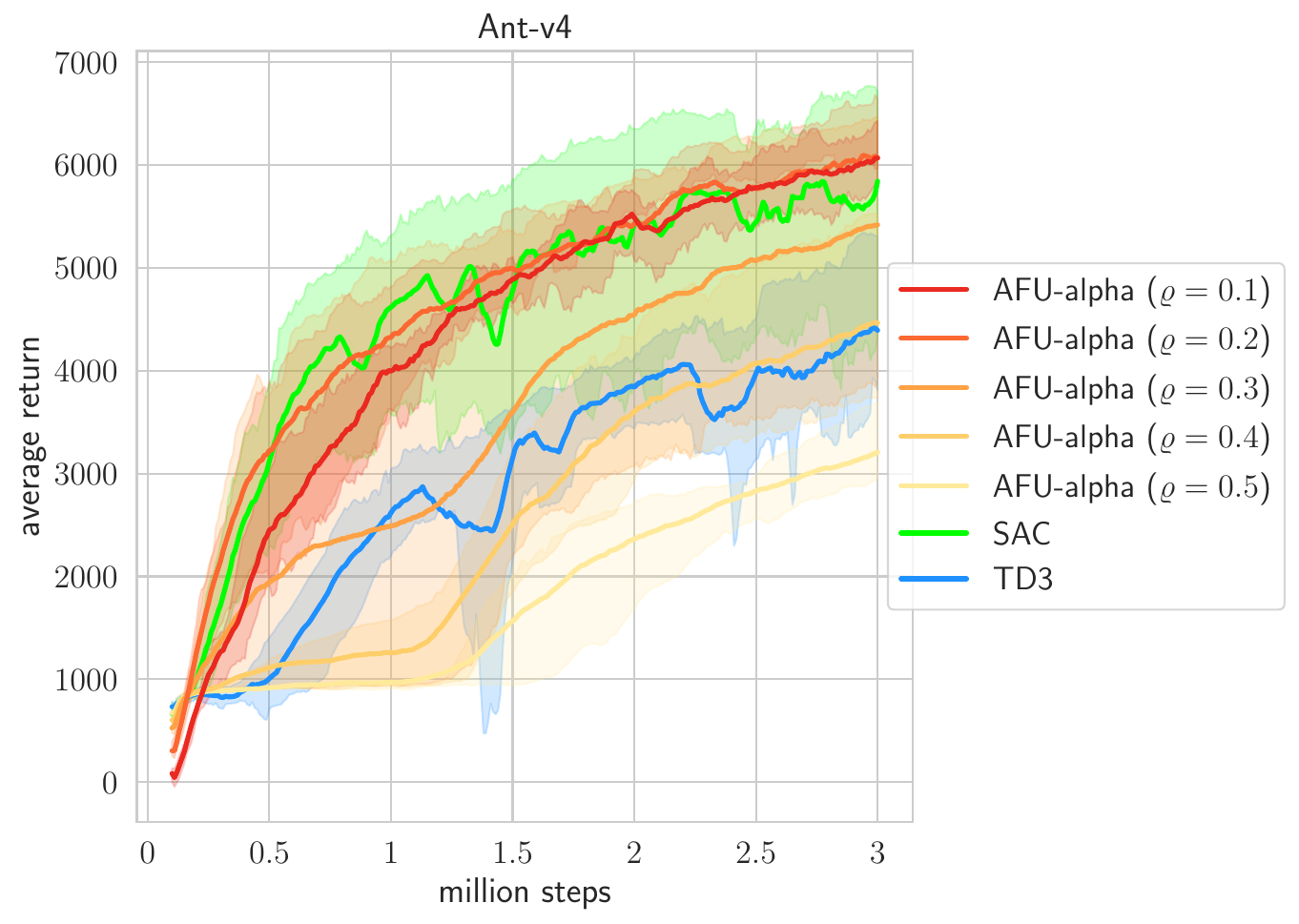}
        \label{fig:ant_alpha}
    \end{subfigure}
    \hspace*{\fill}
    \begin{subfigure}{0.5\linewidth}
        \includegraphics[width=\linewidth]{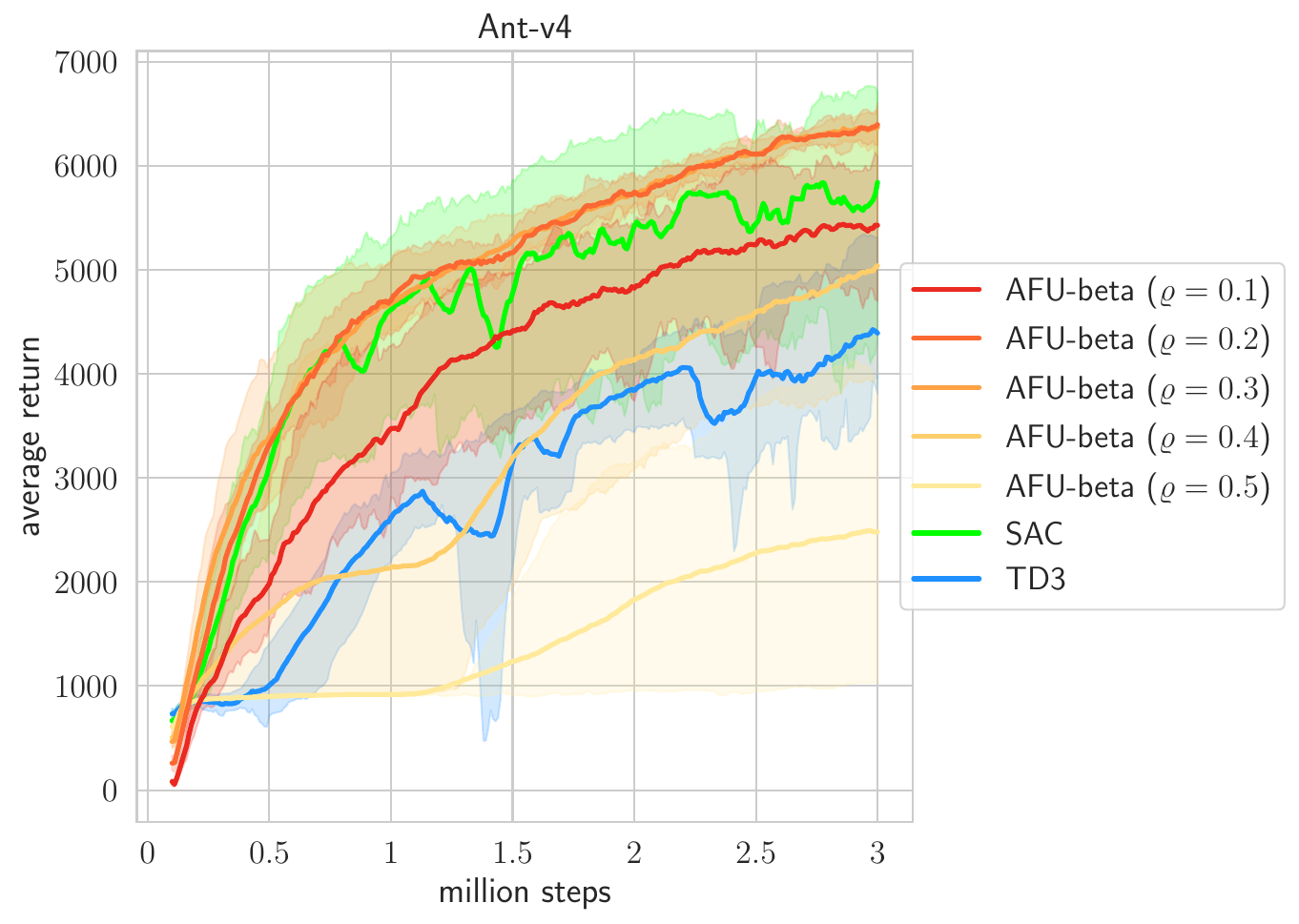}
        \label{fig:ant_beta}
    \end{subfigure}
    \caption{Left: AFU-alpha on Ant-v4. Right: AFU-beta on Ant-v4.}
    \label{fig:ant}
\end{figure}

\begin{figure}[H]
    \begin{subfigure}{0.5\linewidth}
        \includegraphics[width=\linewidth]{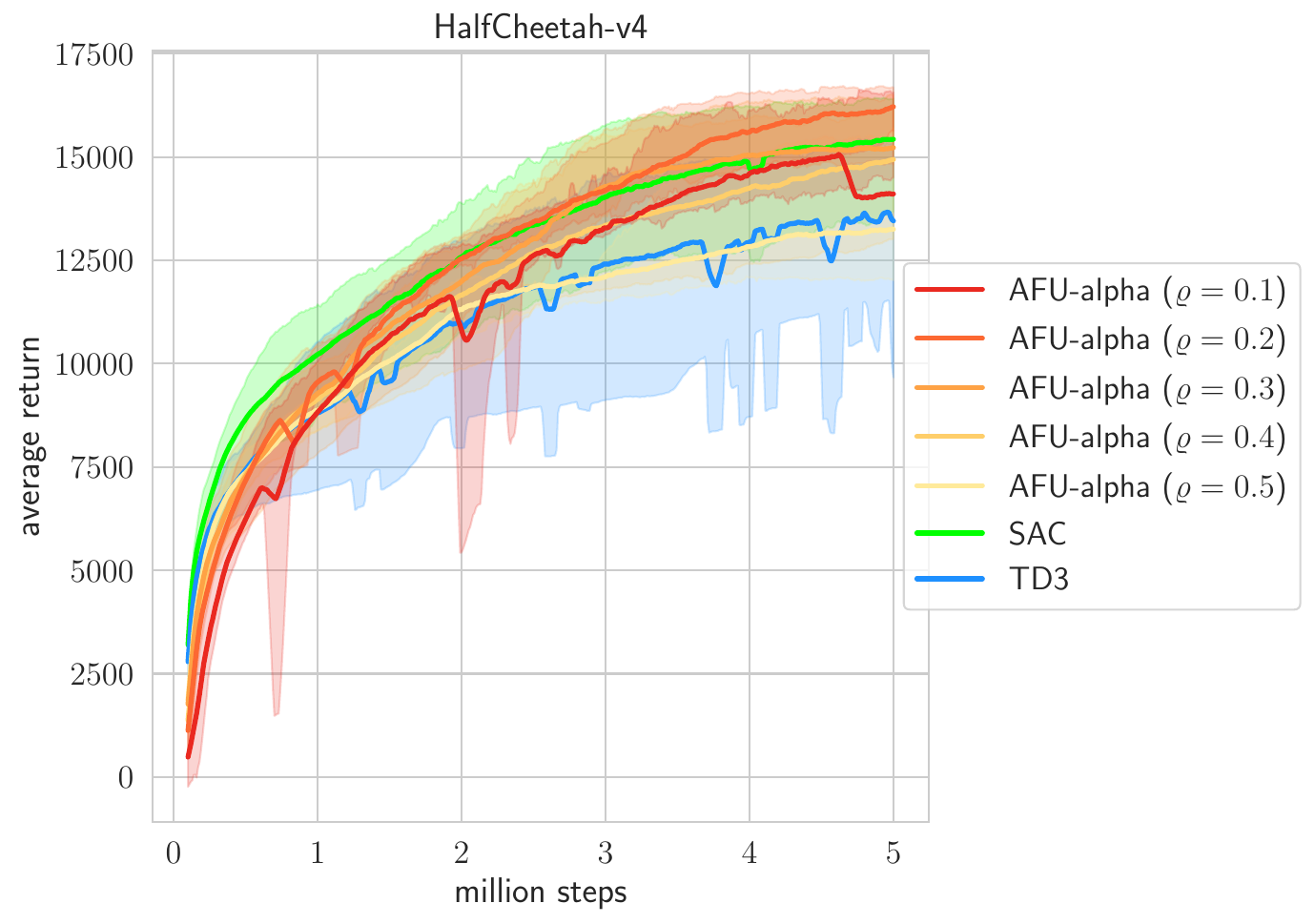}
        \label{fig:hc_alpha}
    \end{subfigure}
    \hspace*{\fill}
    \begin{subfigure}{0.5\linewidth}
        \includegraphics[width=\linewidth]{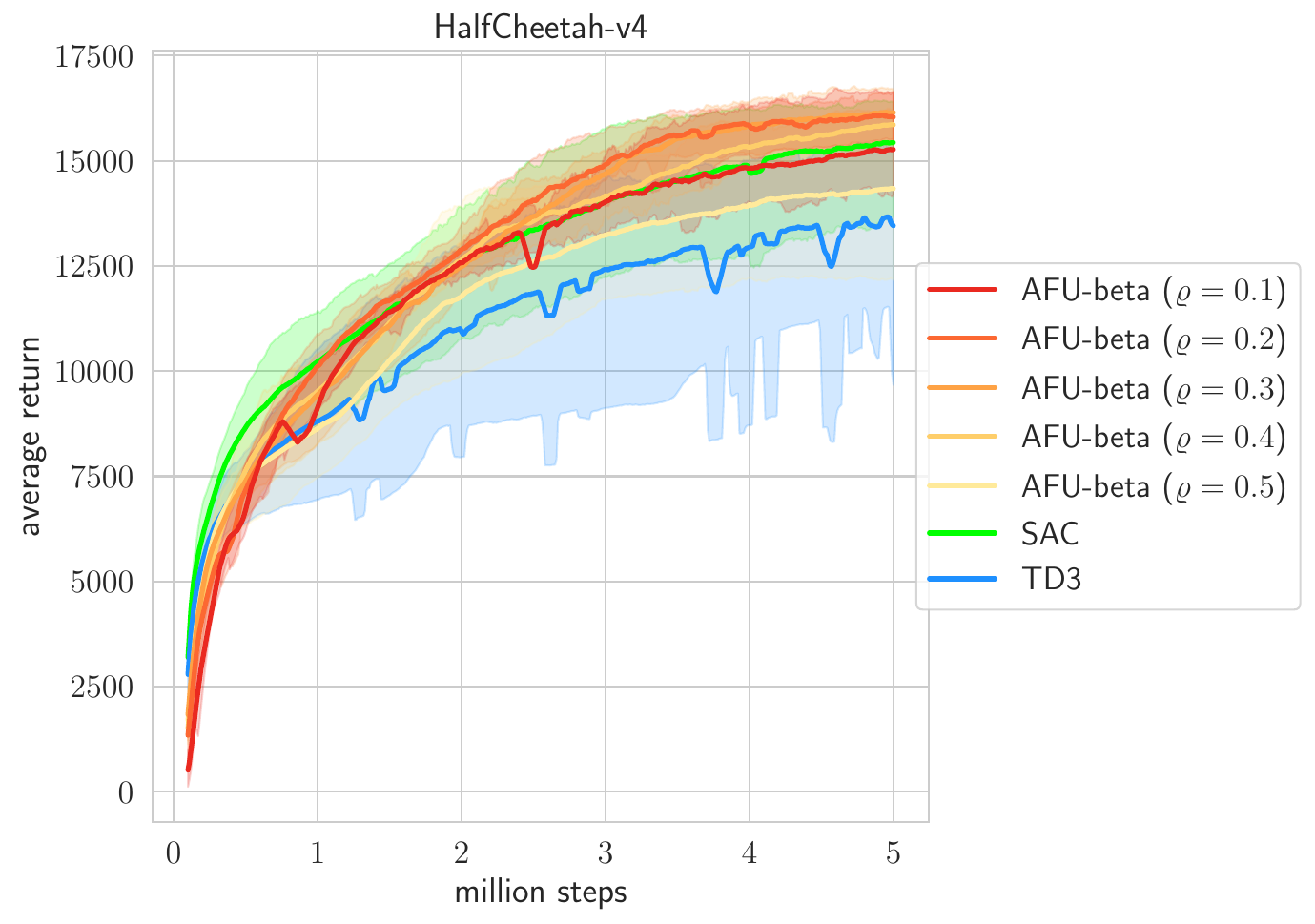}
        \label{fig:hc_beta}
    \end{subfigure}
    \caption{Left: AFU-alpha on HalfCheetah-v4. Right: AFU-beta on HalfCheetah-v4.}
    \label{fig:hc}
\end{figure}

\begin{figure}[H]
    \begin{subfigure}{0.5\linewidth}
        \includegraphics[width=\linewidth]{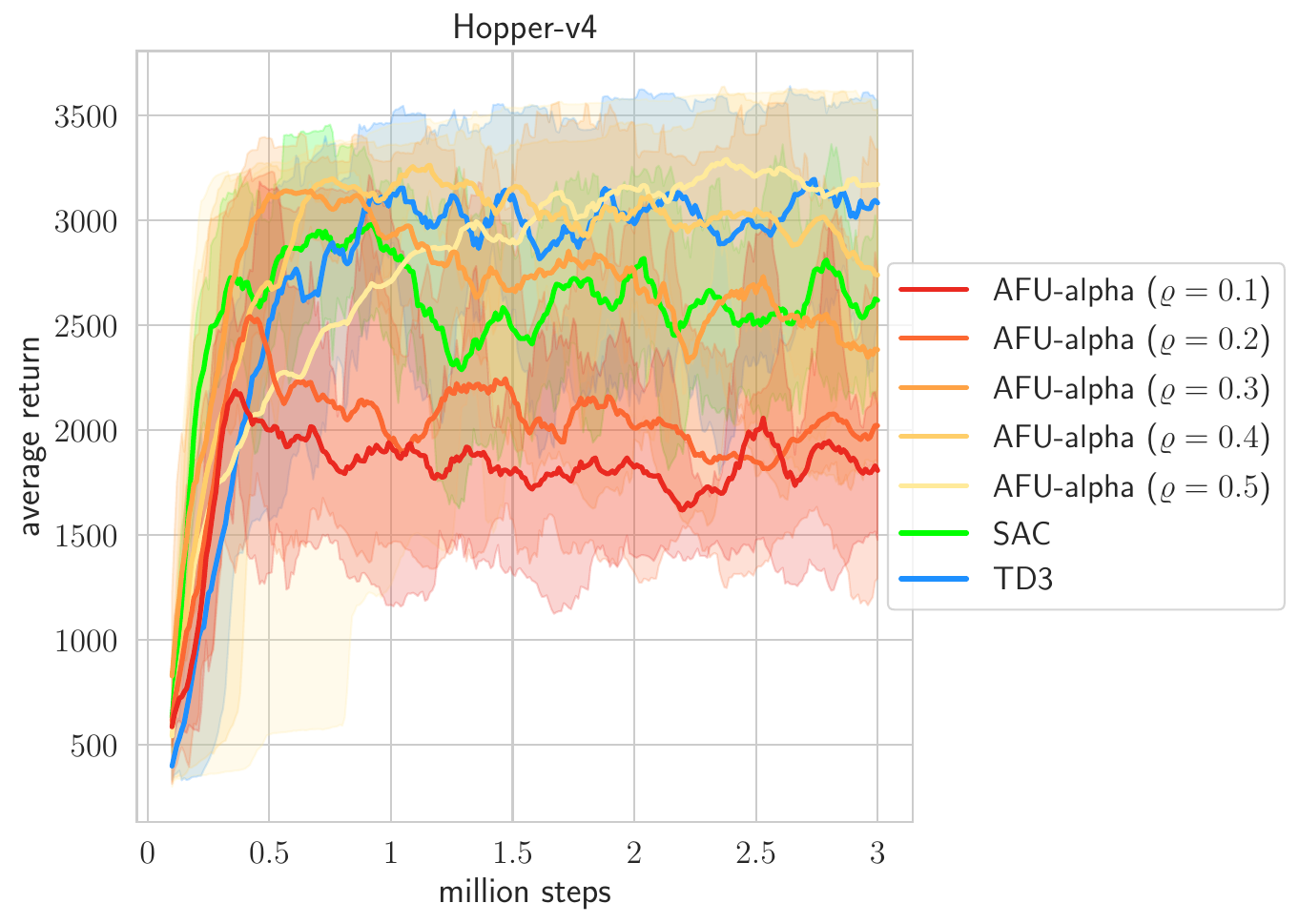}
        \label{fig:hopper_alpha}
    \end{subfigure}
    \hspace*{\fill}
    \begin{subfigure}{0.5\linewidth}
        \includegraphics[width=\linewidth]{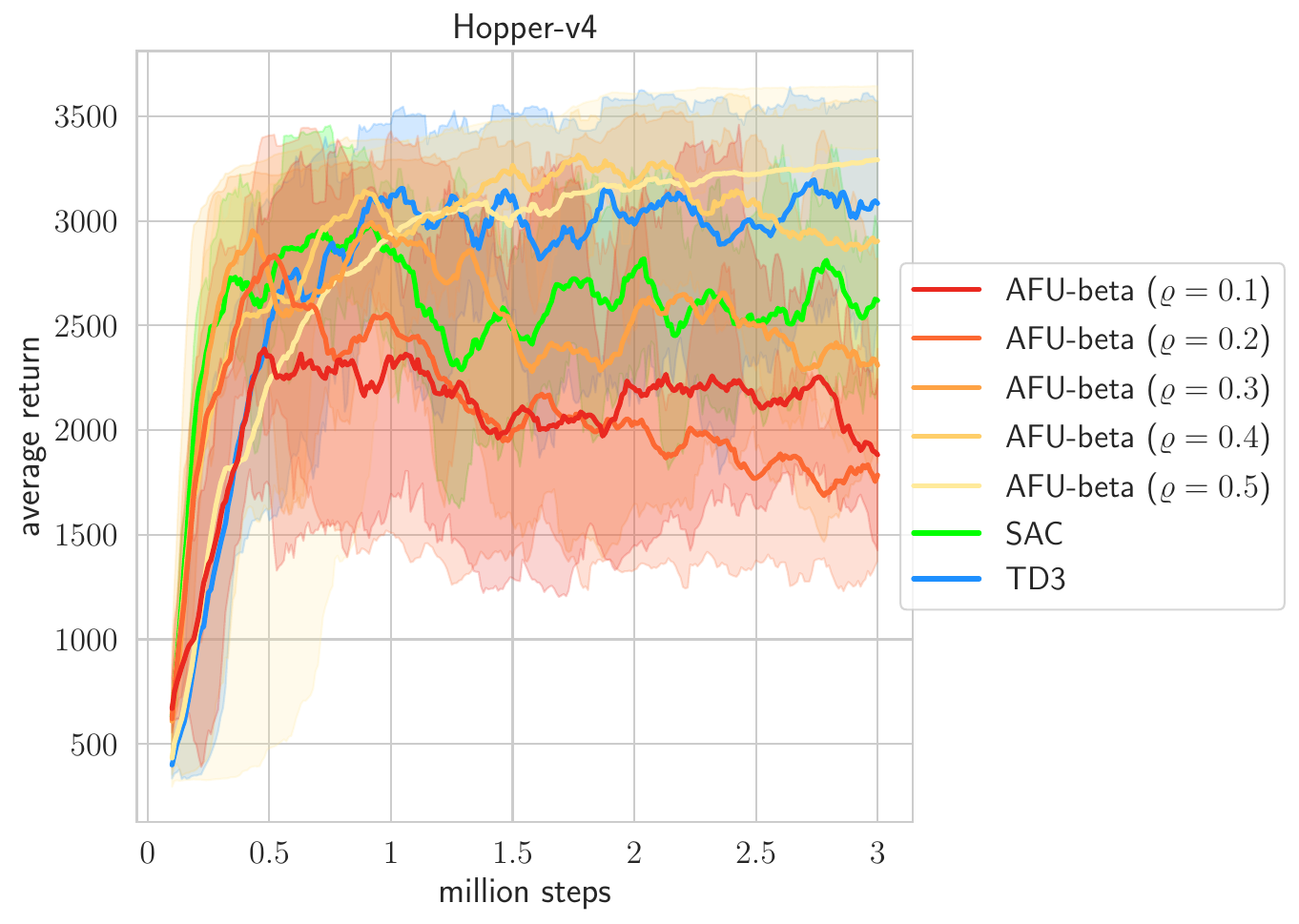}
        \label{fig:hopper_beta}
    \end{subfigure}
    \caption{Left: AFU-alpha on Hopper-v4. Right: AFU-beta on Hopper-v4.}
    \label{fig:hopper}
\end{figure}

\begin{figure}[H]
    \begin{subfigure}{0.5\linewidth}
        \includegraphics[width=\linewidth]{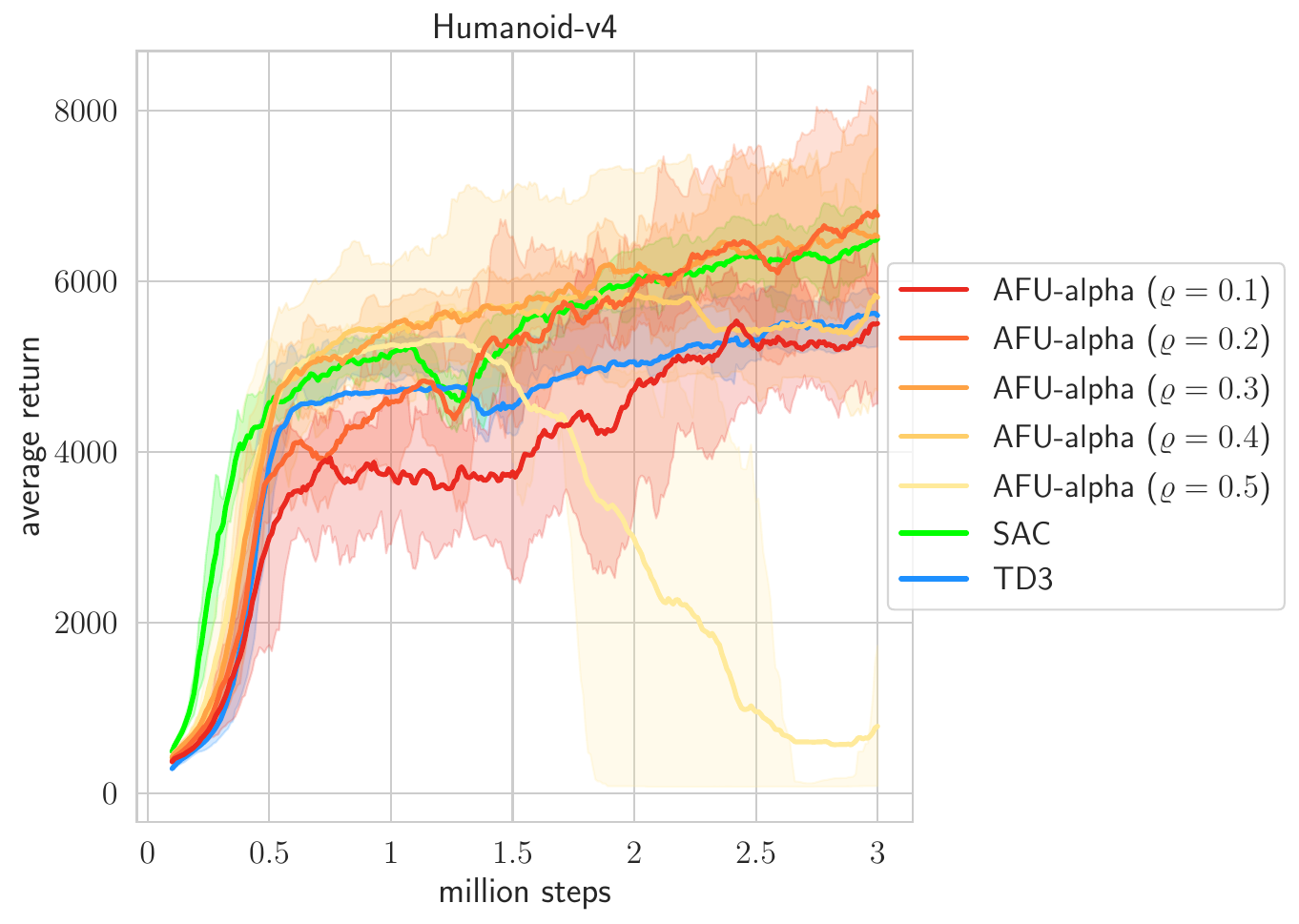}
        \label{fig:humanoid_alpha}
    \end{subfigure}
    \hspace*{\fill}
    \begin{subfigure}{0.5\linewidth}
        \includegraphics[width=\linewidth]{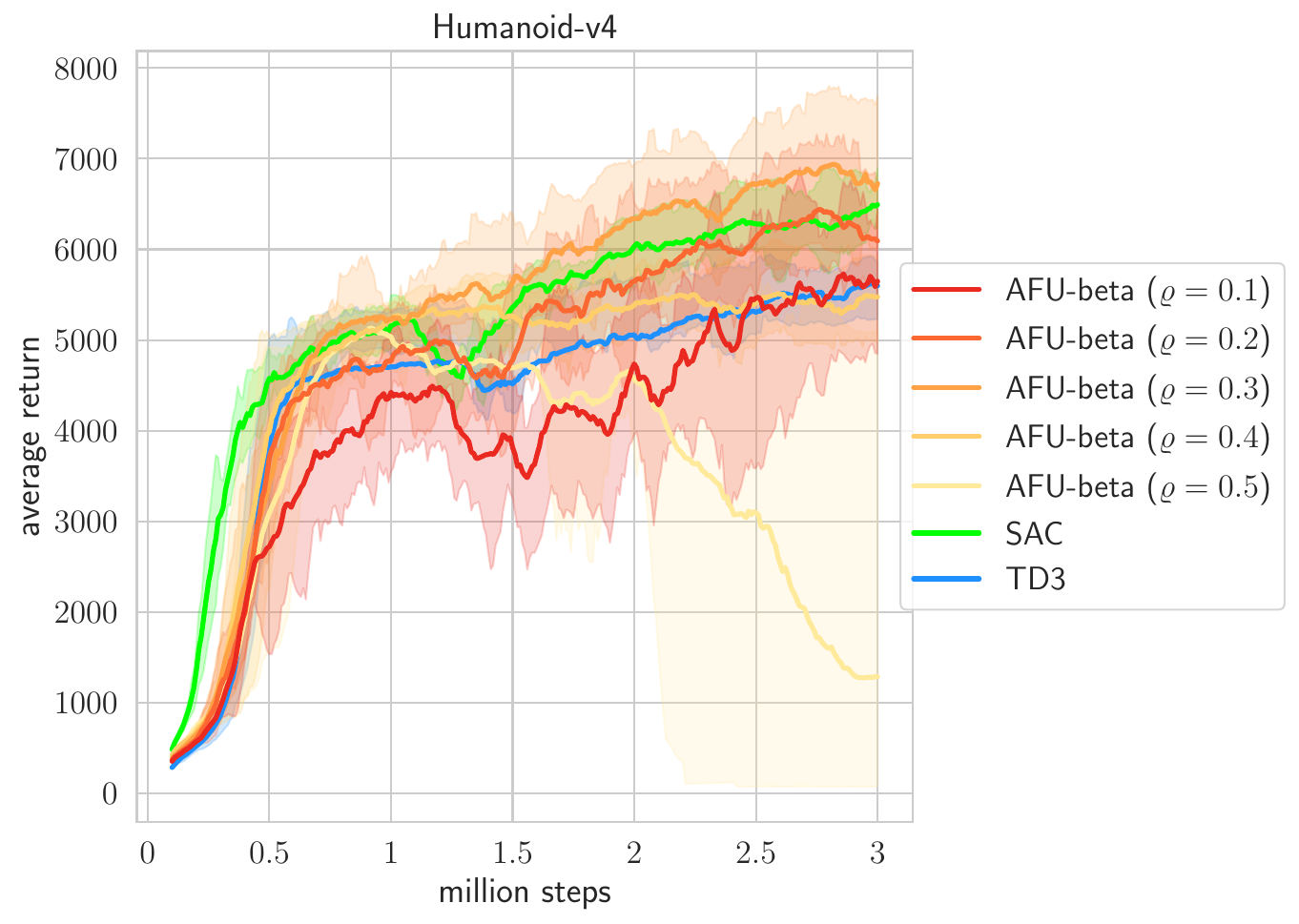}
        \label{fig:humanoid_beta}
    \end{subfigure}
    \caption{Left: AFU-alpha on Humanoid-v4. Right: AFU-beta on Humanoid-v4.}
    \label{fig:humanoid}
\end{figure}

\begin{figure}[H]
    \begin{subfigure}{0.5\linewidth}
        \includegraphics[width=\linewidth]{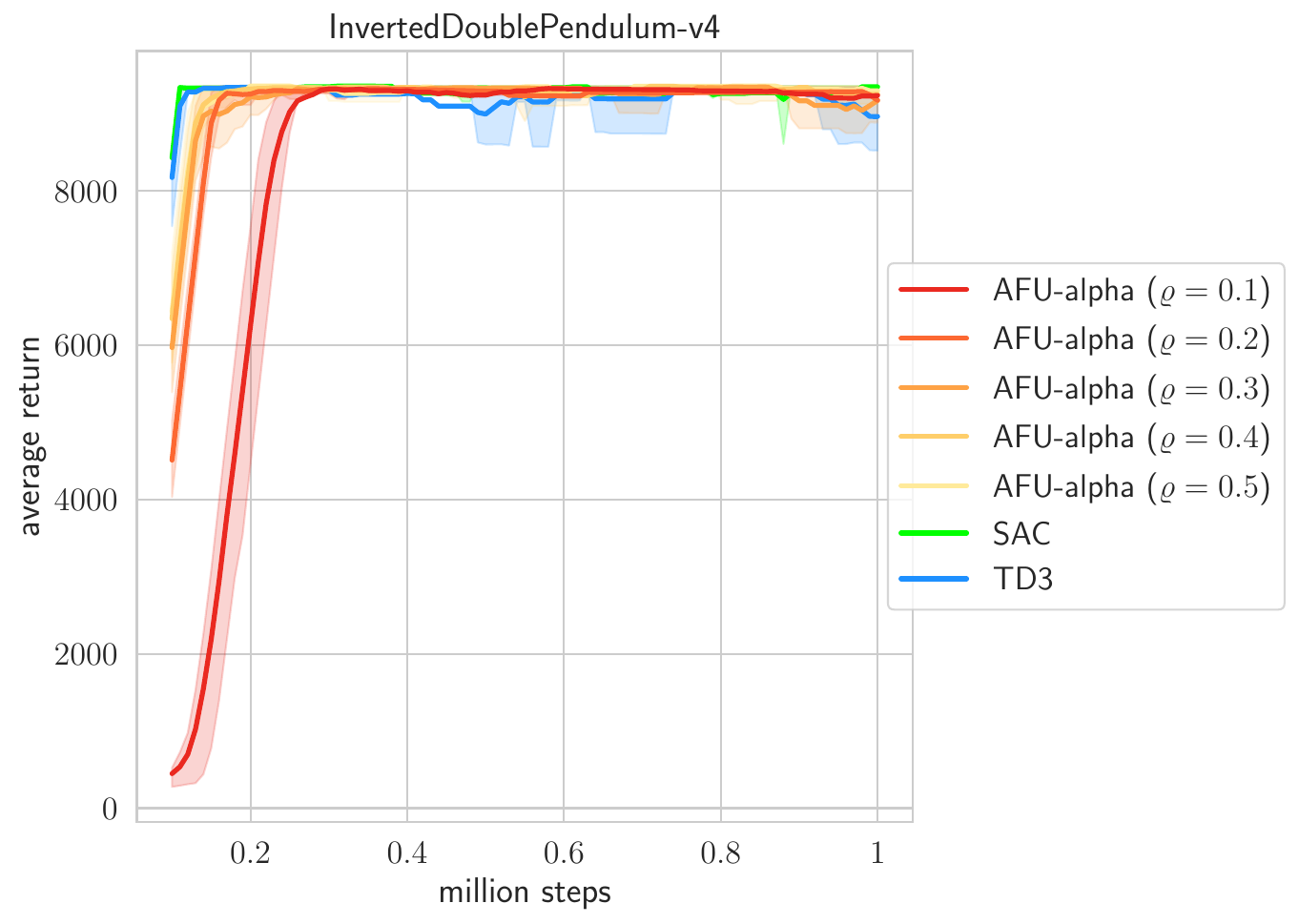}
        \label{fig:idp_alpha}
    \end{subfigure}
    \hspace*{\fill}
    \begin{subfigure}{0.5\linewidth}
        \includegraphics[width=\linewidth]{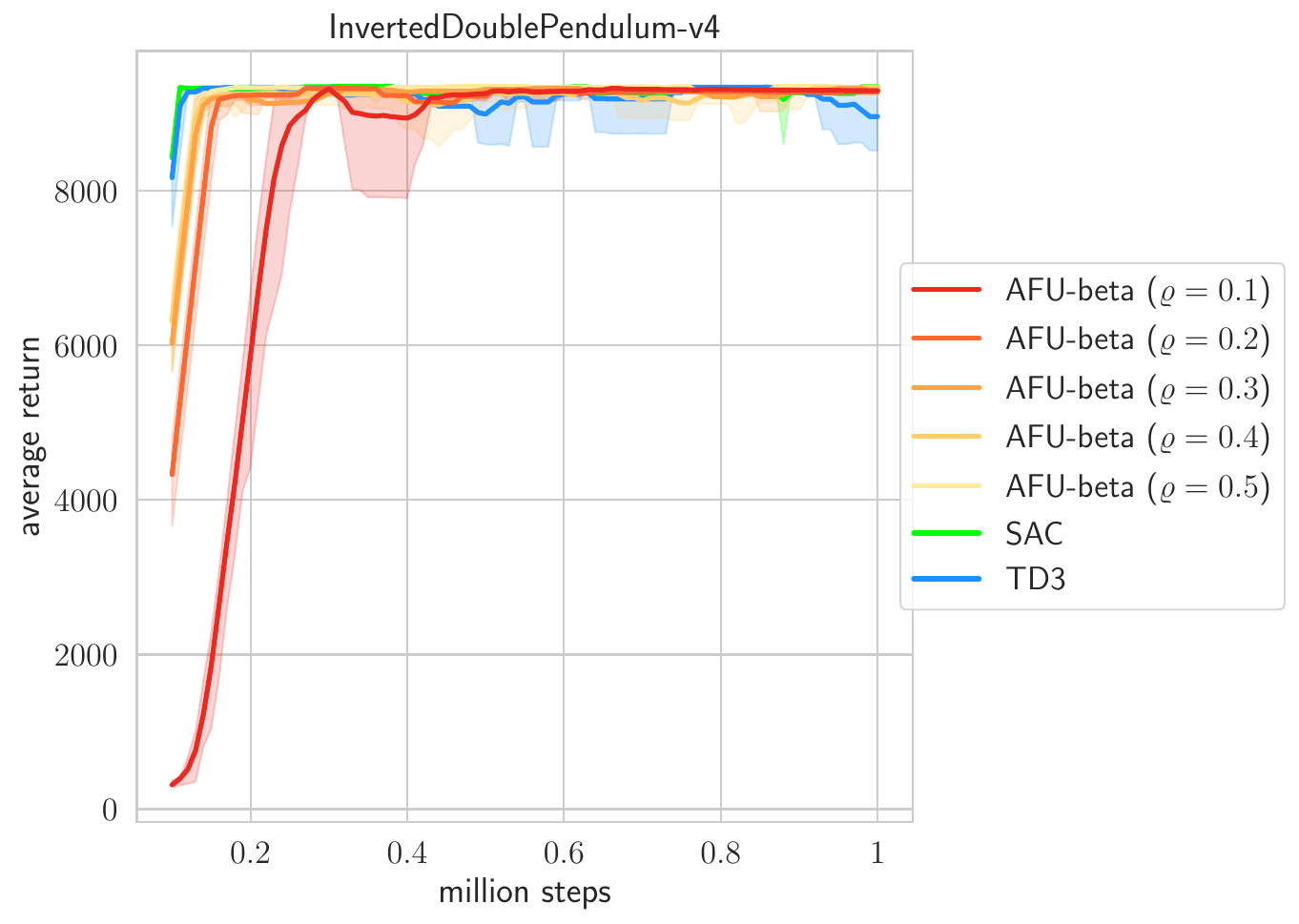}
        \label{fig:idp_beta}
    \end{subfigure}
    \caption{Left: AFU-alpha on InvertedDoublePendulum-v4. Right: AFU-beta on InvertedDoublePendulum-v4.}
    \label{fig:idp}
\end{figure}

\begin{figure}[H]
    \begin{subfigure}{0.5\linewidth}
        \includegraphics[width=\linewidth]{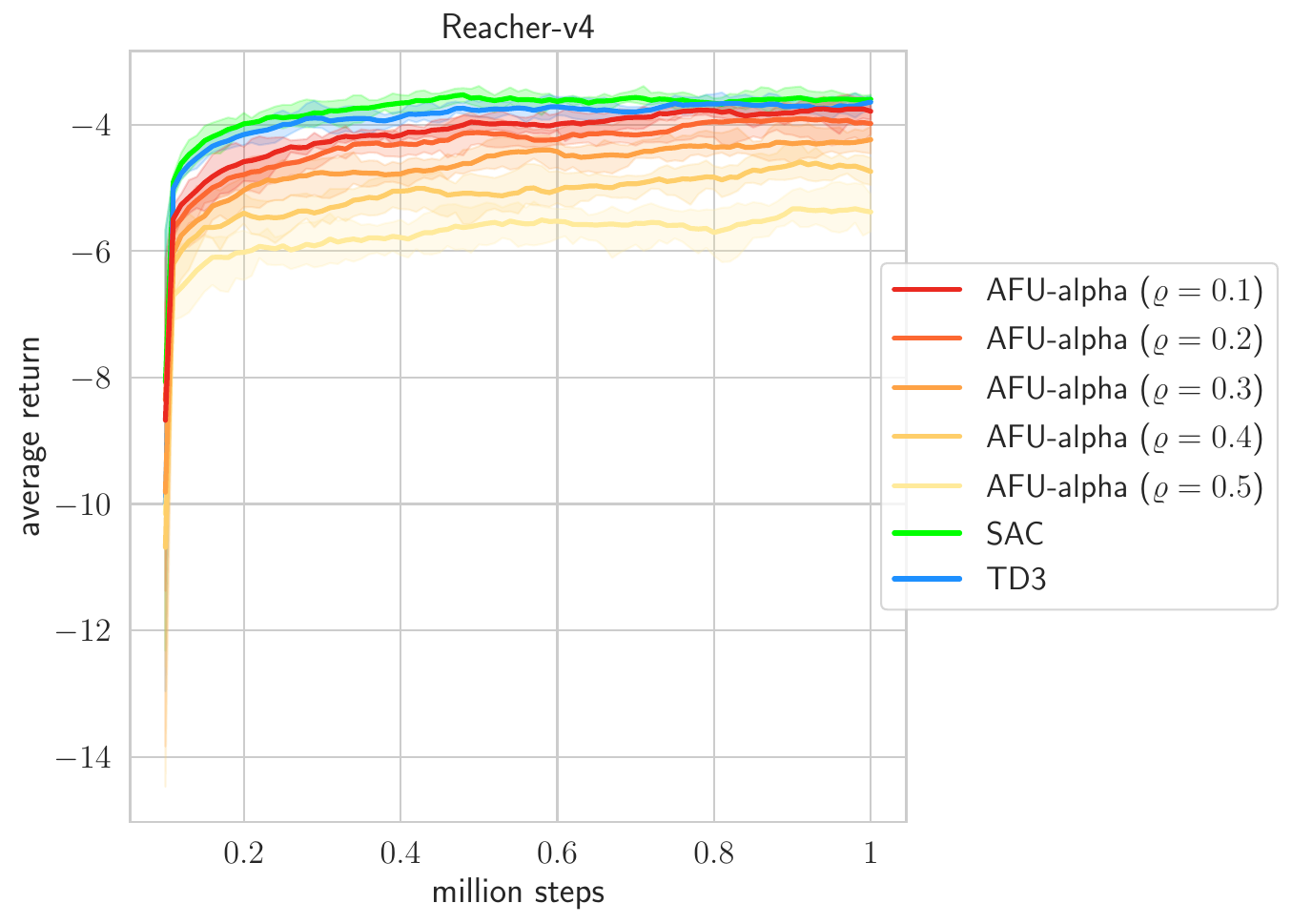}
        \label{fig:reacher_alpha}
    \end{subfigure}
    \hspace*{\fill}
    \begin{subfigure}{0.5\linewidth}
        \includegraphics[width=\linewidth]{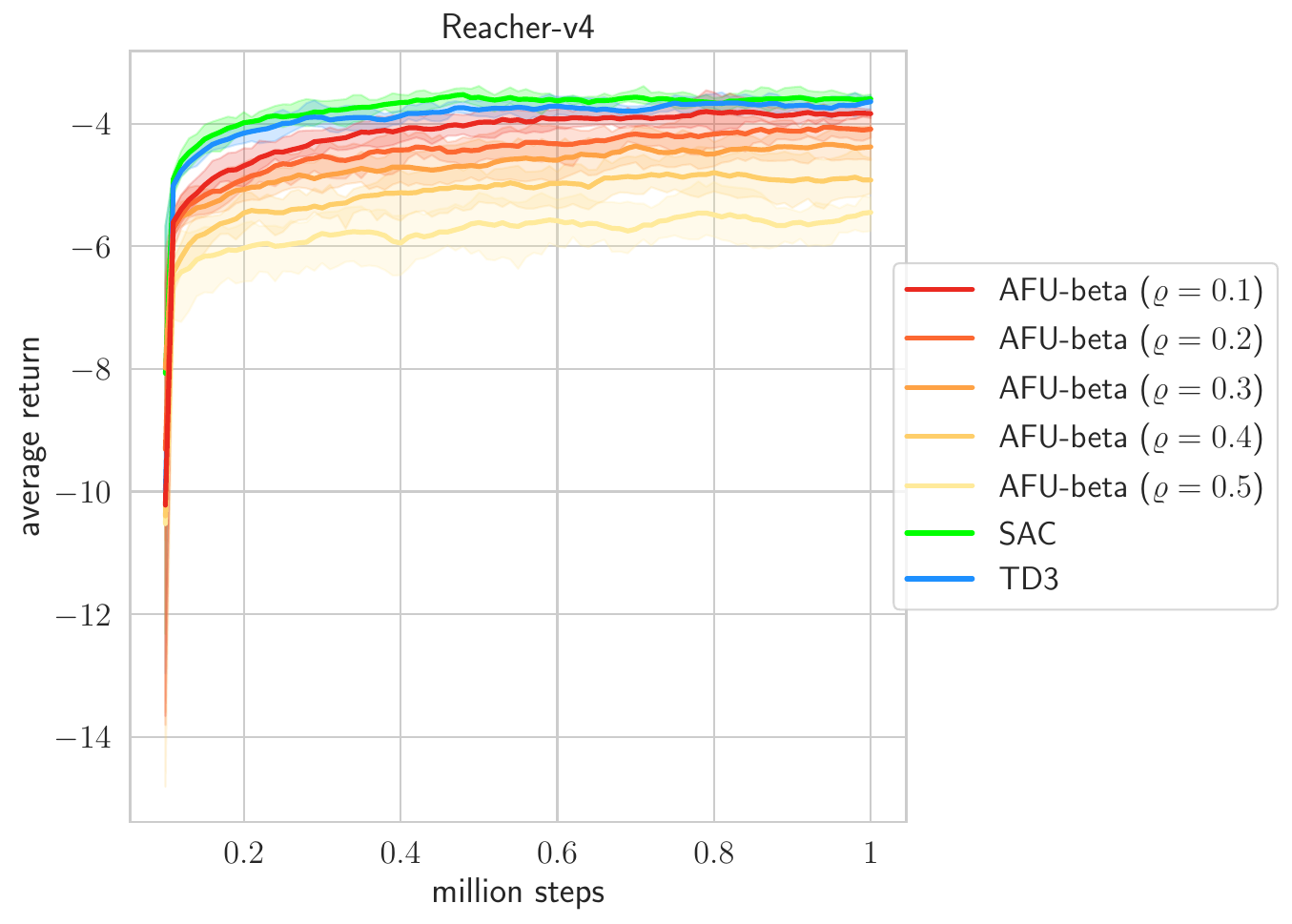}
        \label{fig:reacher_beta}
    \end{subfigure}
    \caption{Left: AFU-alpha on Reacher-v4. Right: AFU-beta on Reacher-v4.}
    \label{fig:reacher}
\end{figure}

\begin{figure}[H]
    \begin{subfigure}{0.5\linewidth}
        \includegraphics[width=\linewidth]{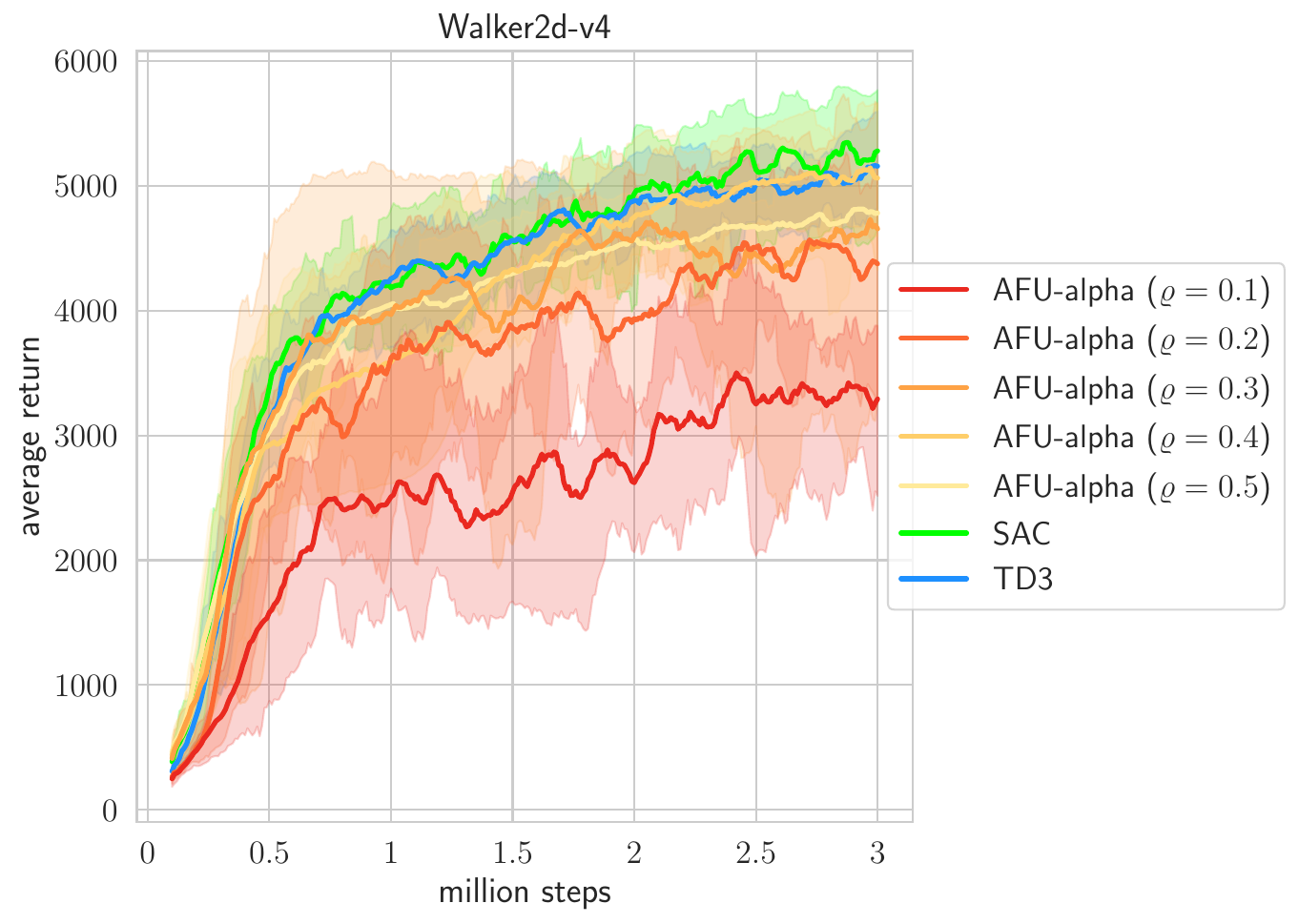}
        \label{fig:walker_alpha}
    \end{subfigure}
    \hspace*{\fill}
    \begin{subfigure}{0.5\linewidth}
        \includegraphics[width=\linewidth]{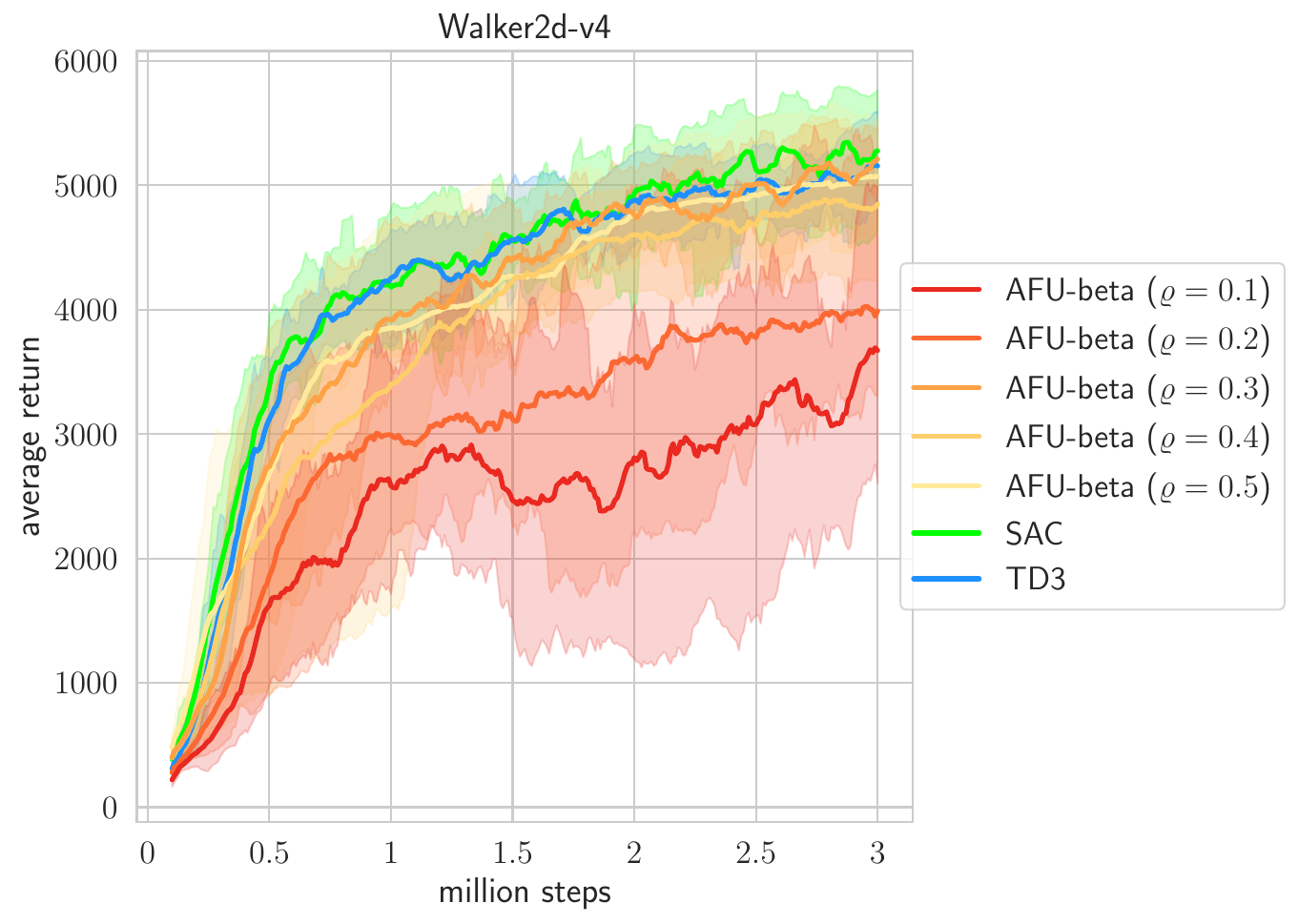}
        \label{fig:walker_beta}
    \end{subfigure}
    \caption{Left: AFU-alpha on Walker2d-v4. Right: AFU-beta on Walker2d-v4.}
    \label{fig:walker}
\end{figure}

\end{document}

%% file: intro.tex
%!TEX root = main.tex

Q-learning \citep{watkins1989learning} stands as a fundamental algorithm in the realm of model-free RL. As mentioned in \cite{watkins1992q}, ``it provides agents with the capability of learning to act optimally in Markovian domains by experiencing the consequences of actions, without requiring them to build maps of the domains''. It is centered around the Bellman optimality equation and leverages dynamic programming to compute or approximate a function known as the optimal  
Q-function $Q^*$. The integration of deep neural networks to approximate Q-functions and the efficient computation of gradient-based updates has led to the successful development of the Deep Q-Network (DQN) algorithm \citep{mnih2015human}, catalyzing important advancements in reinforcement learning. However, Q-learning requires computing the maximum of the Q-function over the action space, which can be difficult if it is continuous and multi-dimensional. To circumvent this ``max-Q problem'', Q-learning can be combined with an actor-critic perspective, which involves coupling policy gradient with Q-learning updates to estimate the action-value function. Although this type of coupling primarily aims 
to evaluate the actor with a Q-learning critic, its byproduct is that the actor is trained to generate actions that maximize the Q-function, thereby solving the max-Q problem indirectly. This approach gave rise to DDPG \citep{lillicrap2015continuous}, a seminal actor-critic algorithm benefiting from the off-policy nature of Q-learning and consequently from a high sample-efficiency. Yet, in DDPG and its derivatives like TD3 \citep{fujimoto2018addressing}, the actor may become trapped in local optima \citep{matheron2020understanding}. Other approaches have attempted to face the max-Q problem head-on, like CAQL \citep{DBLP:conf/iclr/RyuCATB20} or Implicit Q-Learning (IQL, \cite{DBLP:conf/iclr/KostrikovNL22}), but the former does not scale well to high-dimensional action spaces, and requires adaptations such as constraining the action range for complex problems, whereas in the latter, the expectile loss becomes unbalanced when trying to produce estimates that are very close to the true maxima, which has so far restricted the application of IQL and similar methods to offline RL.

Overall, in continous state and action spaces, the most successful modern off-policy deep reinforcement learning algorithms are actor-critic algorithms that tend 
to fail in batch settings, when a large part of the training data is uncorrelated to the distribution under the current policy \cite{DBLP:conf/icml/FujimotoMP19}. In this sense, they 
are not \emph{truly} off-policy. On the other end, truly off-policy algorithms adapted from Q-learning \citep{watkins1989learning}, such as 
Implicit Q-Learning \cite{DBLP:conf/iclr/KostrikovNL22}, are 
adapted to offline RL but do not perform well in online RL. This motivates the quest for a truly off-policy algorithm that is well suited for online RL.
In this paper, we make a step in this direction by proposing a novel way to solve the max-Q problem, using regression and conditional gradient scaling (see Section~\ref{sec:maxq}), resulting in a new algorithm that adapts Q-learning to continuous action spaces. The algorithm still has an actor to select actions and produce episodes, but unlike state-of-the-art model-free off-policy algorithms, most of which are derived from TD3 or SAC \citep{haarnoja2018soft}, its critic updates are entirely independent from the actor. We call the algorithm AFU for ``Actor-Free Updates''. In its first version, AFU-alpha (see Sections \ref{sec:actor} and \ref{sec:alpha}), we use a stochastic actor and train it like the actor in SAC. We then study in Section~\ref{sec:failure} a simple failure mode of SAC (and AFU-alpha), and show that the value function trained by regression in AFU can help improve the actor update and make it less prone to local optima, resulting in a new version of the algorithm, AFU-beta (see Section~\ref{sec:beta}), which does not fail in the same way. 
Our experiments show that AFU-alpha and AFU-beta are competitive in sample-efficiency with TD3 and SAC without being more computationally expensive. To the best of our knowledge, AFU is the first model-free off-policy RL algorithm that is competitive with the state-of-the-art and truly departs from the actor-critic perspective. 
%Besides, by reducing the risk of being ensnared in local optima, we believe that AFU-beta represents an advancement in improving the off-policy nature  of RL algorithms dedicated to continuous control problems.

%% file: rw.tex
In domains where high sample efficiency is crucial, such as robotics, the off-policy nature of RL algorithms becomes paramount. This allows training on samples obtained from different policies or older versions of the current policy, facilitating faster learning and compatibility with various exploration strategies.
One way to obtain an off-policy algorithm is to adapt Q-learning, but as previously mentioned, in continuous action spaces, direct approaches attempting to solve the max-Q problem of Q-learning have faced limitations. Besides CAQL, which formulates the max-Q problem as a mixed-integer program, and IQL, which treats Q-functions as state-dependent random variables and relies on expectile regression to estimate their maxima, we can cite NAF \citep{gu2016continuousnaf} and ICNN \citep{amos2017input}, which impose action-convex Q-functions making the max-Q problem tractable, QT-Opt \citep{kalashnikov2018qt}, which uses a stochastic optimizer to tackle non-convex max-Q problems, or approaches based on a discretization of the action space, such as SMC-learning \citep{lazaric2007reinforcement} and SDQN \citep{metz2017discrete}. However, these methods often struggle with complex, high-dimensional continuous control tasks, either due to a lack of expressiveness or prohibitive computational costs.
Close to IQL, $\mathcal{X}$-QL \citep{DBLP:conf/iclr/GargHGE23} is an offline RL algorithm relying on an objective directly estimating the optimal soft-value function in the maximum entropy RL setting without needing to sample from a policy. A variant of $\mathcal{X}$-QL \citep{DBLP:conf/iclr/GargHGE23} works in the online setting, but in this case critic updates depend on actions sampled by the actor for the Bellman backup. A unique off-policy algorithm, AWR \citep{peng2019advantage}, employs regression to train a value function and a policy but falls short of the sample efficiency achieved by state-of-the-art off-policy algorithms. 
Presently, the most successful approaches in model-free off-policy RL for continuous control are actor-critic algorithms with interwoven actor and critic updates. 
The first off-policy actor-critic algorithm was introduced in \cite{degris2012off}, and the most recent ones are typically based on TD3, an improvement of DDPG, or on SAC, which relies on an entropy maximization framework that led to various off-policy algorithms by creating connections between policy gradients and Q-learning updates (see \cite{o2016combining}).
 Among the algorithms improving upon TD3 and SAC, we can mention TQC \citep{kuznetsov2020controlling}, a distributional approach to control the overestimation bias, REDQ \citep{chen2021randomized} or AQE \citep{wu2021aggressive} which employ critic ensembles, DroQ \citep{hiraoka2021dropout} which uses dropout and layer normalization in the critic networks, and BAC \citep{ji2023seizing} which merges Q-function updates from SAC and IQL. While these ideas could be incorporated into our proposed algorithm AFU, we leave this for future work and focus on comparing AFU to SAC and TD3. In contrast to methods building upon SAC and TD3, AFU is structurally distinct because the critic updates remain unaffected by the actor. Notably, the critic is never trained with out-of-distribution (OOD) actions, yet AFU achieves a level of sample efficiency competitive with SAC and TD3. One might object that OOD actions can be beneficial in the online setting, because they favor exploration. As pointed out in \cite{DBLP:conf/iclr/GargHGE23}, OOD actions in Bellman backups introduce over-optimism, but online learning allows agents to correct over-optimism by collecting additional data. Yet, by achieving results comparable in sample-efficiency to TD3 and SAC without OOD actions, we show that the benefit of Bellman backups with OOD actions in the online setting is in fact not so obvious. If OOD actions can introduce an over-optimism that then needs to be corrected, it may be preferable to design online learning methods that do not yield over-optimism in the first place, and use other strategies to favor exploration. Furthermore, unlike other direct adaptations of Q-learning to continuous control, AFU does not fail on the most complex tasks. On the contrary, the challenging MuJoCo task Humanoid is one of the environments in which AFU performs the best comparatively to SAC and TD3.

%% file: prelim.tex
We consider a discounted infinite horizon Markov Decision Problem (MDP) $<S, A, T, R, \gamma>$, where $S$ is a state space, $A$ a continuous action space, $T$ a stochastic transition function, $R : S \times A \rightarrow \mathbb{R}$ a reward function, and $0 \leq \gamma < 1$ a discount factor. 
We denote by $s'$ (resp. $s_{t+1}$) a state obtained after performing an action $a$ (resp. $a_t$) in state $s$ (resp. $s_t$). Transitions are tuples $(s, a, r, s')$ with $r = R(s,a)$. The optimal Q-function $Q^*$ is defined by:
$
Q^*(s, a) = \mathbb{E}\left[\sum_{t=0}^{\infty} \gamma^t R(s_t, a_t) \mid s_0 = s, a_0 = a, \pi^* \right],
$
where the policy used from $t=1$ onwards is $\pi^*$, which selects actions optimally in every state. 
The optimal value function $V^*$ verifies
$
V^*(s) = \max_{a \in A} (Q^*(s, a)).
$
Let $V_{\varphi_1}$ and $V_{\varphi_2}$ denote two function approximators for the value function, and $Q_{\psi}$ a function approximator for the Q-function (the critic). 
We use feed forward neural networks for all the function approximators. For the value function, we also consider target networks (see \cite{mnih2016asynchronous}), i.e. parameter vectors $\varphi^\text{target}_1$ and $\varphi^\text{target}_2$ updated with the rule \mbox{$\varphi^\text{target}_i \leftarrow \tau\varphi_i + (1-\tau) \varphi^\text{target}_i$} for some target smoothing coefficient $0 < \tau < 1$.
We wish to train the critic on mini-batches $B$ of transitions $(s, a, r, s')$ taken from an experience replay buffer, with the following loss derived from the clipped Double Q-learning loss of TD3 \citep{fujimoto2018addressing}:
\begin{align}
\label{eq:LQ}
L_Q(\psi) = \underset{(s, a, r, s') \in B}{\text{Mean}}\left[ \left( Q_\psi(s, a) - r - \gamma \min_{i \in \{1, 2\}}V_{\varphi^\text{target}_i}(s') \right)^2 \right]
\end{align}

The use of two function approximators $V_{\varphi_1}$ and $V_{\varphi_2}$ aims at avoiding the overestimation bias that can make Q-learning based approaches diverge (see \cite{hasselt2010double}). In practice, transitions can be terminal, which requires a simple modification of the loss ignored here for the sake of clarity ($\gamma$ is set to $0$ for terminal transitions).
Provided that $V_{\varphi_1}(s)$ and $V_{\varphi_2}(s)$ return good estimates of the maximum of $Q_\psi(s, \cdot)$, a maximization usually referred to as the max-Q problem (see \cite{DBLP:conf/iclr/RyuCATB20}), Equation~\eqref{eq:LQ} amounts to the mean squared Bellman error that drives $Q_\psi$ toward $Q^*$ \citep{baird1999reinforcement}.

%% file: maxq.tex
The main remaining problem is: how to efficiently train $V_{\varphi_1}$ and $V_{\varphi_2}$? For the learning to be successful, $V_{\varphi_1}$ and $V_{\varphi_2}$ should both converge to precise solutions of the max-Q problem, and the convergence should be fast, because if changes in $Q_\psi$ are not tracked promptly, errors such as overestimation of Q-values could lead to failures.

\subsection{Method}

We introduce two new function approximators $A_{\xi_1}$ and $A_{\xi_2}$, for the optimal advantage function defined by
$
A^*(s,a) = Q^*(s,a) - V^*(s).
$
For any state-action pair $(s, a)$, $A^*(s,a)\leq 0$. Preliminarily, we assume that outputs of $A_{\xi_i}$ can only be non-positive. Assuming that $Q_\psi$ is fixed, training $V_{\varphi_i}$ and $A_{\xi_i}$ can be done by minimizing the following regression loss on mini-batches $B$: 
\begin{align}
\label{eq:lVAinit}
l_{V, A}(\varphi_i, \xi_i) = \underset{(s, a, \_, \_) \in B}{\text{Mean}}\left[ \Big( V_{\varphi_i}(s) + A_{\xi_i}(s,a) - Q_\psi(s, a) \Big)^2 \right].
\end{align}
This loss causes the values $V_{\varphi_i}(s)$ to become upper bounds of $Q_\psi(s, \cdot)$, but not tight ones. 
A natural next step would be to add a regularization term penalizing large outputs of $V_{\varphi_i}$, which results in an approach very similar to methods based on regression with asymmetric losses such as IQL, SQL, EQL and $\mathcal{X}$-QL \citep{DBLP:conf/iclr/GargHGE23}. The issue is that the resulting convergence is either slow (for very small regularization coefficients) or significantly biased (for larger coefficients). For some problems, finding the right coefficient is possible, but in general standard regularization does not lead to satisfactory results in the context of online RL. We propose a different approach based on conditional gradient rescaling, noticing that when $V_{\varphi_i}(s) + A_{\xi_i}(s,a)$ is greater than its target $Q_\psi(s, a)$, by gradient descent both $V_{\varphi_i}(s)$ and $A_{\xi_i}(s,a)$ would decrease by the same amount, and conversely when $V_{\varphi_i}(s) + A_{\xi_i}(s,a)$ is smaller than the target, values would both increase by the same amount. Without regularization, all upper bounds of $Q_\psi(s, \cdot)$ are equally good values for $V_{\varphi_i}(s)$, but we can asymmetrically modulate the gradients to put a ``downward pressure'' on $V_{\varphi_i}(s)$ and make it progressively decrease as $A_{\xi_i}(s,a)$ progressively increases. To this end, we apply only a fraction of the gradient descent update on $\varphi_i$ when $V_{\varphi_i}(s)$ would increase. It can be done by defining, for $0 < \varrho < 1$:
$$
\Upsilon_{i}^{a}(s) = \left(1 - \varrho I_{i}^{s,a} \right) V_{\varphi_i}(s) + \varrho I_{i}^{s,a} V_{\varphi^{\text{no\_grad}}_i}(s),
$$
where $\varphi^{\text{no\_grad}}_i$ is a copy of the parameters $\varphi_i$, and $
I_{i}^{s,a} =\left\{
    \begin{array}{ll}
      1, & \mbox{if $V_{\varphi_i}(s) + A_{\xi_i}(s,a) < Q_\psi(s, a)$}.\\
      0, & \mbox{otherwise}.
    \end{array}
  \right.
$
Replacing $V_{\varphi_i}(s)$ by $\Upsilon_{i}^{a}(s)$ in \eqref{eq:lVAinit} yields a new version of the loss:
\begin{align}
\label{eq:LambdaVA}
\Lambda_{V, A}(\varphi_i, \xi_i) = \underset{(s, a, \_, \_) \in B}{\text{Mean}}\left[ \Bigl(\Upsilon_{i}^{a}(s) + A_{\xi_i}(s,a) - Q_\psi(s, a) \Bigr)^2 \right].
\end{align}
Remark: the proposed method based on conditional gradient rescaling is similar to an adaptive regularization scheme in which the weight of the regularization is proportional to the absolute value of the error, see Appendix~\ref{app:rescaling}. 

So far, we have assumed that $A_{\xi_i}(s,a)$, typically the output of a neural network, can only be non-positive. 
But imposing a strict constraint on the sign of $A_{\xi_i}(s,a)$ could potentially lead to jittering gradients, so we instead restrict its sign in a soft way (see Appendix~\ref{app:soft}), resulting in this loss:
\begin{align}
\label{eq:LambdaVAmodif}
\Lambda'_{V, A}(\varphi_i, \xi_i) = \underset{(s, a, \_, \_) \in B}{\text{Mean}}\left[  Z\Bigl(\Upsilon_{i}^{a}(s) - Q_\psi(s, a), A_{\xi_i}(s,a) \Bigr) \right],
\end{align}
with $Z(x,y) =\left\{
    \begin{array}{l}
      (x+y)^2 \mbox{ if $x \geq 0$}.\\
      x^2+y^2 \mbox{ otherwise}.
    \end{array}
  \right.$

\subsection{Experiments}

\begin{figure}[htbp]
    \begin{subfigure}{0.5\linewidth}
        \includegraphics[width=\linewidth]{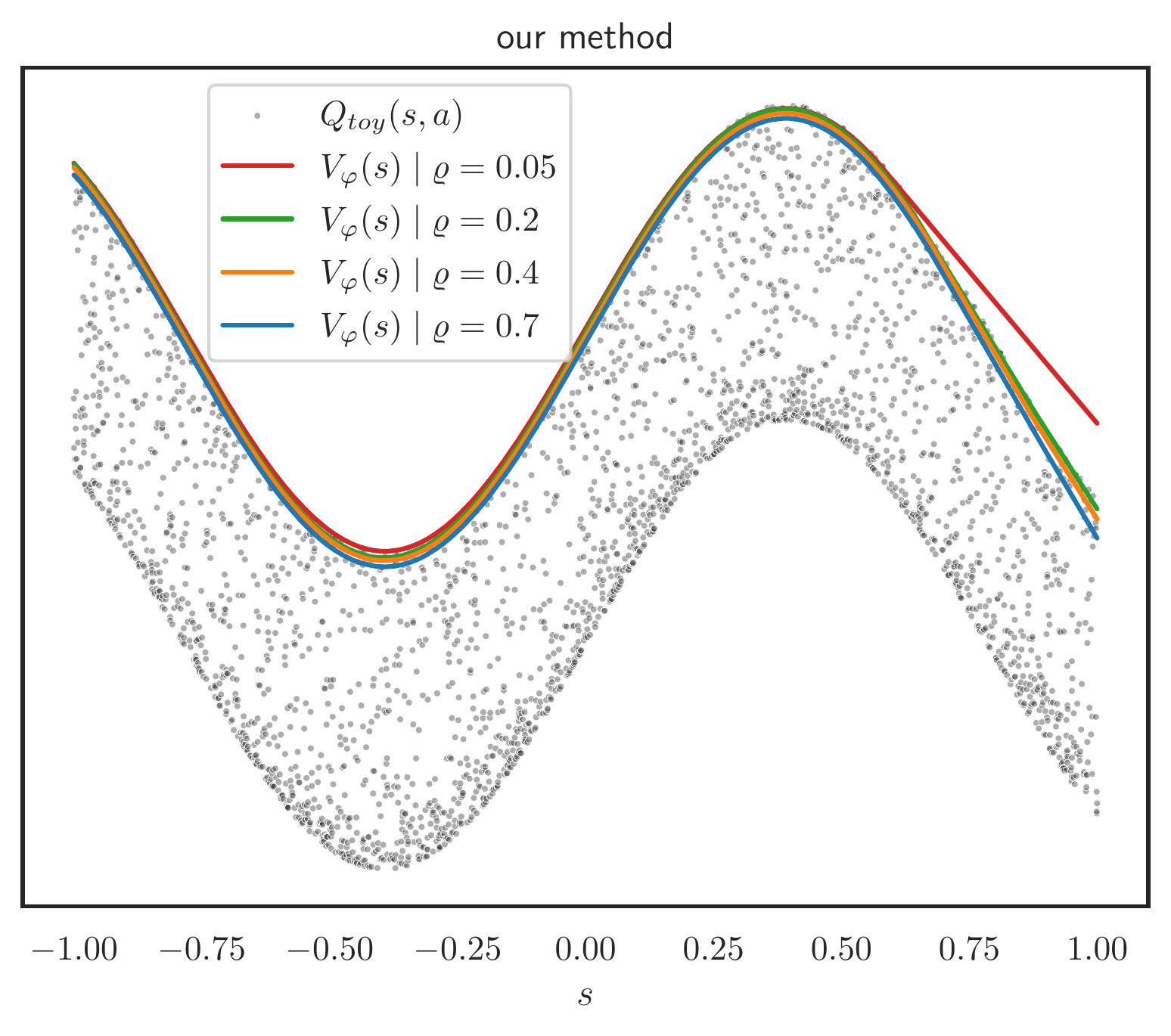}
        \caption{$V_{\varphi}(s)$ is trained jointly with $A_{\xi}(s,a)$ by iterating gradient descent steps on the loss $\Lambda'_{V, A}(\varphi, \xi)$ described by Equation~\eqref{eq:LambdaVAmodif}. $\varrho \in [0.2, 0.7]$ results in precise approximations of $s \mapsto \max_{a \in A} (Q_\psi(s, a))$.}
        \label{fig:toyproblem_sub1_main}
    \end{subfigure}
    \hspace*{\fill}
    \begin{subfigure}{0.5\linewidth}
        \includegraphics[width=\linewidth]{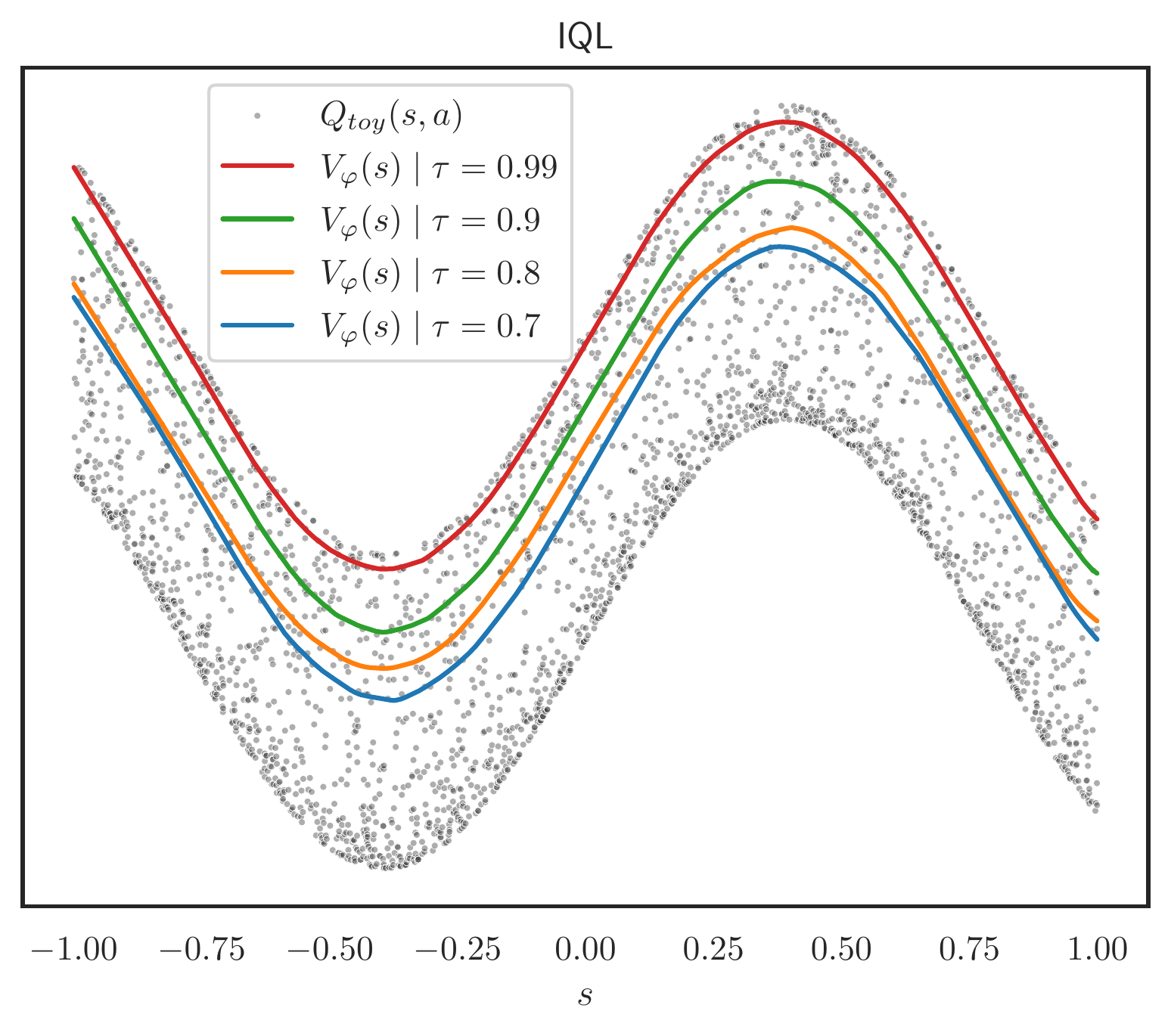}
        \caption{Results of the training with the loss from IQL \citep{DBLP:conf/iclr/KostrikovNL22} for 4 different values of the hyperparameter $\tau$. Values used in actual (offline) RL experiments are not greater than $0.9$.}
        \label{fig:toyproblem_sub2_main}
    \end{subfigure}
    \caption{$Q_{toy}(s,a) = \sin(4s) + 0.7\cos(4a)$
for $(s, a) \in [-1, 1]^2$. Our method and IQL both train $V_{\varphi}(s)$ to approximate $s \mapsto \max_{a \in A} (Q_{toy}(s, a))$, i.e. solve a max-Q problem. Trainings are done with 3000 gradient descent steps on batches of 256 uniformly randomly drawn values of $(s, a)$.}
    \label{fig:toyproblem}
\end{figure}

We empirically compare our method to 3 baselines on a toy problem.
We define the function $Q_{toy}(s,a) = \sin(4s) + 0.7\cos(4a)$
for $s \in [-1, 1]$ and $a \in [-1, 1]$.
We use a single feedforward neural network for $V$ ($V_\varphi$) and a single feedforward neural network for $A$ ($A_\xi$). Both networks have two hidden layers of size 256 and ReLU activations in the hidden layers. Our method trains both $V_\varphi$ and $A_\xi$, while the 3 baselines IQL, SQL and EQL directly train $V_\varphi$. All 3 baselines have been successfully applied to offline reinforcement learning.
IQL is Implicit Q-Learning \citep{DBLP:conf/iclr/KostrikovNL22}, and SQL and EQL are respectively Sparse Q-Learning and Exponential Q-Learning, both introduced in \cite{DBLP:conf/iclr/Xu0LYWCZ23}. For a fixed $s$, IQL treats $Q_{toy}(s,a)$ as a random variable (the randomness being determined by the action) and uses an expectile regression loss to train $V_\varphi(s)$ to estimate a state conditional upper expectile of this random variable. The expectile is determined by the parameter $0 < \tau < 1$, and the closer $\tau$ is to 1, the closer $V_\varphi(s)$ gets to $\max_{a \in A} (Q_{toy}(s, a))$. 
However, if $\tau$ is very close to $1$ (e.g. $\tau = 0.99$), the loss becomes unbalanced, with elements weighted hundreds of times more than others, which results in instabilities in the context of RL, so in practice the values used in \cite{DBLP:conf/iclr/KostrikovNL22} are not greater than $0.9$. Figure~\ref{fig:toyproblem} compares our method to IQL. We observe that, with our method, although $\varrho=0.05$ leads to overestimations, a wide range of parameter values (from $\varrho=0.2$ to $\varrho=0.7$) yield precise results, while with IQL a precise result is only obtained with a hyperparameter value inapplicable to RL ($\tau = 0.99$). In Appendix~\ref{app:toy}, the more complete Figure~\ref{fig:toyproblem_appendix} shows a comparison to IQL, SQL and EQL. The same observation can be made for all 3 baselines: results with hyperparameters that are suitable to RL are significantly less precise than the ones obtained with our method.

%% file: actor.tex
First, we remove the dependency to $Q_\psi$ in loss \eqref{eq:LambdaVAmodif} by replacing $Q_\psi(s, a)$ by the targets used to train it in \eqref{eq:LQ}. We obtain the following loss (for $i \in \{1, 2\}$):
\begin{align}
\label{eq:LVA}
L_{V, A}(\varphi_i, \xi_i) = \underset{(s, a, r, s') \in B}{\text{Mean}}\left[  Z\Bigl(\Upsilon_{i}^{a}(s) -  r - \gamma \min_{i \in \{1, 2\}}V_{\varphi^\text{target}_i}(s'), A_{\xi_i}(s,a) \Bigr) \right].
\end{align}
With the losses $L_Q(\psi)$ \eqref{eq:LQ} and $L_{V, A}(\varphi_i, \xi_i)$ \eqref{eq:LVA}, we can train $Q_\psi$, $V_{\varphi_i}$ and $A_{\xi_i}$ without needing an actor.
Compared to methods derived from DDPG (like TD3), solving directly the max-Q problem has an advantage over first using an actor to solve the argmax-Q problem, i.e. to approximate $\text{argmax}_{a \in A} (Q_\psi(s, a))$. The reason is that continuous changes in $Q_\psi$ result in continuous changes of its state conditioned maxima, while it can result in discontinuous changes of its state conditioned argmax. So, in an off-policy setting, if the exploration policy discovers better results with very different actions, the maximum of $Q_\psi(s,a)$ can be tracked smoothly, while the tracking of the argmax can be much more difficult, with the potential arising of deceptive value landscapes in which the actor can get stuck (see \cite{matheron2020understanding}). This theoretical advantage, as well as the
actor-free $Q_\psi$, $V_{\varphi_i}$ and $A_{\xi_i}$ updates are all important aspects of our approach. However, since we are interested in online reinforcement learning, we still need an actor to select actions and produce episodes. To train this actor, if we would use the same gradient ascent over $a \mapsto Q_\psi(s,a)$ as in DDPG, our global method would be prone to the same failure modes as DDPG, and most of the advantages of the max-Q based training of $V_{\varphi_i}$ would be lost.
One thing we can notice is that, since we do not need the actor to return 
$\text{argmax}_{a \in A} (Q_\psi(s, a))$, we also do not need the actor to be deterministic.
To benefit from a better exploration, we opt for a stochastic actor and follow the approach proposed in SAC \citep{haarnoja2018soft} with automatic tuning of the temperature parameter $\alpha$. It relies on two losses $L_{\pi}(\theta)$ and $L_{\text{temp}}(\alpha)$, and on a target entropy $\bar{\mathcal{H}}$ (see Appendix~\ref{app:sac}):
\begin{align}
\label{eq:Lpi}
L_{\pi}(\theta) = \underset{\substack{(s, \_, \_, \_) \in B \\ a_s \sim \pi_\theta(\cdot | s)}}{\text{Mean}}\Big[ \alpha \log(\pi_\theta(a_s | s)) - Q_\psi(s, a_s) \Big].
\end{align}
\begin{align}
\label{eq:Ltemp}
L_{\text{temp}}(\alpha) = \underset{\substack{(s, \_, \_, \_) \in B \\ a_s \sim \pi_\theta(\cdot | s)}}{\text{Mean}}\Big[ -\alpha \log(\pi_\theta(a_s | s)) - \alpha \bar{\mathcal{H}} \Big].
\end{align}

%% file: afualpha.tex
We combine the losses $L_Q$ \eqref{eq:LQ}, $L_{V,A}$ \eqref{eq:LVA}, $L_\pi$ \eqref{eq:Lpi}, and $L_{\text{temp}}$ \eqref{eq:Ltemp} to devise a new off-policy reinforcement learning algorithm. 
It has a critic ($Q_\psi$) and an actor ($\pi_\theta$), but the critic updates are derived from our novel adaptation of Q-learning to continuous action spaces (obtained with our new method to solve the max-Q problem), therefore they are independent from the actor. Hence the name AFU for the algorithm, for ``Actor-Free Updates''. 
We specifically call it AFU-alpha to contrast it with AFU-beta introduced in Section~\ref{sec:beta}. AFU-alpha, described in Algorithm~\ref{alg:afualpha}, alternates between environments steps that gather experience in a replay buffer and gradient steps that draw batches from the replay buffer to compute loss gradients and update all parameters of the function approximators. In our implementation, an iteration consists of a single environment step followed by a single gradient step.

\begin{algorithm}[htbp]
\caption{\textcolor{blue}{AFU-alpha} and \textcolor{red}{AFU-beta}}
\label{alg:afualpha}
\begin{algorithmic}
\STATE \mbox{Set $0 < \varrho < 1$, $0 < \tau < 1$, $\bar{\mathcal{H}}$, and learning rates $\eta_Q$, $\eta_{V,A}$, $\eta_\pi$, $\eta_\text{temp}$.}
\STATE \mbox{Initialize empty replay buffer $\mathfrak{R}_b$, and params $\psi$, $\varphi_1 = \varphi^\text{target}_1$, $\varphi_2 = \varphi^\text{target}_2$, $\xi_1$, $\xi_2$, $\alpha$, $\theta$.}
\FOR{each iteration}
	\FOR{each environment step}
	\STATE \mbox{Sample action $a \sim \pi_\theta(\cdot | s)$.}
	\STATE \mbox{Perform environment step $s, a \rightarrow s'$, compute $r = R(s,a)$, and insert $(s, a, r, s')$ in $\mathfrak{R}_b$.}
	\ENDFOR
	\FOR{each gradient step}
	\STATE \mbox{Draw batch of transitions $B$ from $\mathfrak{R}_b$ and compute loss gradients on that batch.}
	\STATE \mbox{$\psi \leftarrow \psi - \eta_Q \nabla_\psi L_Q(\psi)$}
	\STATE \mbox{$\varphi_{i \in \{1,2\}} \leftarrow \varphi_i - \eta_{V, A} \nabla_{\varphi_i} L_{V, A}(\varphi_i, \xi_i)$}
	\STATE \mbox{$\xi_{i \in \{1,2\}} \leftarrow \xi_i - \eta_{V, A} \nabla_{\xi_i} L_{V, A}(\varphi_i, \xi_i)$}
	\STATE \mbox{$\varphi^\text{target}_{i \in \{1,2\}} \leftarrow \tau \varphi_i + (1 - \tau)\varphi^\text{target}_i$}
	\STATE \mbox{\hphantom{$\theta \leftarrow \theta - \eta_{\pi} \nabla_\theta L_{\pi}(\theta)$} \qquad \qquad \qquad \qquad \textcolor{red}{$\zeta \leftarrow \zeta - \eta_{\pi} \nabla_\zeta L_{\mu}(\zeta)$}}
	\STATE \mbox{\textcolor{blue}{$\theta \leftarrow \theta - \eta_{\pi} \nabla_\theta L_{\pi}(\theta)$} \qquad \qquad \qquad \qquad \textcolor{red}{$\theta \leftarrow \theta - \eta_{\pi} \nabla^{\text{MODIF}}_\theta L_{\pi}(\theta)$}}
	\STATE \mbox{$\alpha \leftarrow \alpha - \eta_{\text{temp}} \nabla_\alpha L_{\text{temp}}(\alpha)$}
	\ENDFOR
\ENDFOR
\end{algorithmic}
\end{algorithm}

\begin{figure}[htbp]
    \begin{subfigure}{0.54\linewidth}
        \includegraphics[width=\linewidth]{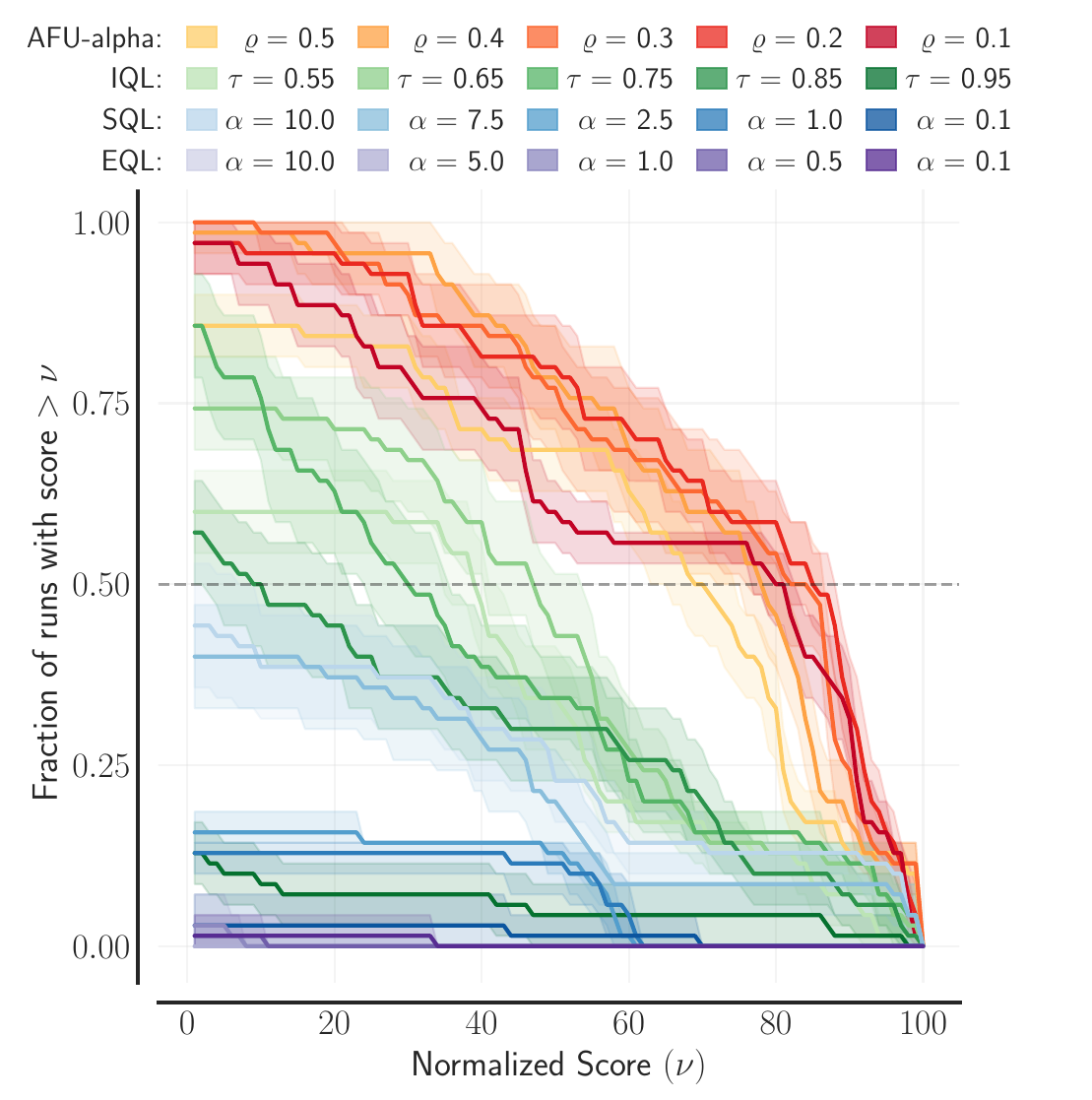}
        \caption{AFU-alpha works best with $\varrho \in \{0.2, 0.3\}$. Using the IQL, SQL and EQL baselines to solve the max-Q problem results in a clear performance deterioration.}
        \label{fig:XP_AFUalpha_sub1}
    \end{subfigure}
    \hspace*{\fill}
    \begin{subfigure}{0.45\linewidth}
        \includegraphics[width=\linewidth]{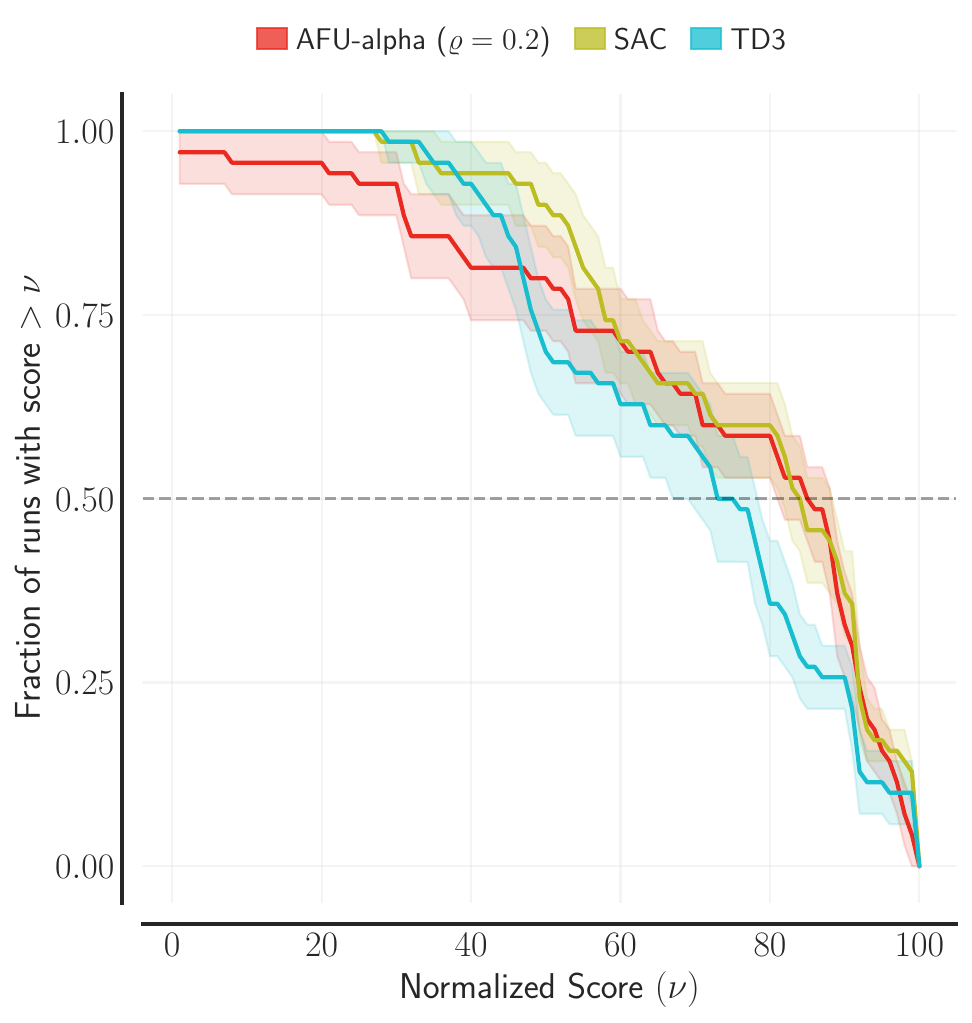}
        \caption{AFU-alpha is competitive with SAC and TD3 on a benchmark of 7 diverse tasks, with action space dimensions ranging from 1 (InvertedDoublePendulum) to 17 (Humanoid), and observation space dimensions ranging from 11 to 376 (Humanoid).}
        \label{fig:XP_AFUalpha_sub2}
    \end{subfigure}
    \caption{Experimental evaluation of AFU-alpha on a benchmark of 7 MuJoCo tasks.}
    \label{fig:XP_AFUalpha}
\end{figure}

\paragraph{Experiments}
We test AFU-alpha on a classical benchmark of 7 MuJoCo \citep{todorov2012mujoco} tasks from the Gymnasium library \citep{towers_gymnasium_2023}. We compare it to SAC and TD3, and to variants of AFU-alpha in which the loss $L_{V,A}$ aiming at solving the max-Q problem is replaced by the corresponding loss taken from IQL, SQL or EQL. The results are shown in Figure~\ref{fig:XP_AFUalpha}.
For both SAC and AFU-alpha, we use the same heuristic for the definition of $\bar{\mathcal{H}}$: we set it to $-d$, where $d$ is the dimension of the action space.
Updates in AFU-alpha, SAC and TD3 use the same value of $\tau$ and same learning rates.
For each algorithm, for each value of the hyperparameter ($\varrho$ for AFU-alpha, $\tau$ for IQL, $\alpha$ for SQL and EQL), and for each of the 7 MuJoCo tasks, we perform 10 runs initialized with different random seeds, and evaluate the performance of the policy every 10,000 steps on 10 rollouts. The first 10,000 steps 
of each run use uniformly drawn random actions (and no gradient steps). Learning curves are smoothed with a moving average window of size 10. The raw score of a run is the last average return, i.e. the average return over the last 10 evaluations.
For each task, we linearly rescale the scores based on two reference points: (1) the maximum evaluation seen across all algorithms and all runs corresponds to a score of 100, and (2) the mean episode return across all algorithms and runs corresponds to a score of 0.
Following the recommendations of \cite{agarwal2021deep}, we compute with the \emph{rliable} library the performance profiles for each algorithm across the 7 tasks: Ant-v4, HalfCheetah-v4, Hopper-v4, Humanoid-v4, InvertedDoublePendulum-v4, Reacher-v4 and Walker2d-v4. The length of the runs is 1 million steps for InvertedDoublePendulum and Reacher, 3 million steps for Ant, Hopper, Humanoid and Walker2d, and 5 million steps for HalfCheetah.

In Figure~\ref{fig:XP_AFUalpha_sub1}, we see that our proposed method for the max-Q problem yields significantly better results than the IQL, SQL and EQL baselines. The best results are obtained with $\varrho \in \{0.2, 0.3\}$.
Figure~\ref{fig:XP_AFUalpha_sub2} shows that AFU-alpha is competitive with SAC and TD3.

%Normalization: reference scores:
%Ant: 1202.3 - 6993.6
%HalfCheetah: 4660.1 - 17411.0
%Hopper: 1300.1 - 4113.0
%Humanoid: 2179.3 - 9734.4
%InvertedDoublePendulum: 5306.9 - 9360.0
%Reacher: -18.16 - -2.27
%Walker2d: 1808.4 - 6324.7
%
%
%Normalization: reference scores (with AFU-beta):
%Ant: 1269.7 - 6993.6
%HalfCheetah: 4795.4 - 17309.5
%Hopper: 1315.4 - 4113.0
%Humanoid: 2181.6 - 8526.8
%InvertedDoublePendulum: 5300.6 - 9360.0
%Reacher: -18.17 - -2.29
%Walker2d: 1777.8 - 6192.7

%% file: failureSAC.tex
\begin{figure}[htbp]
	\includegraphics[width=\linewidth]{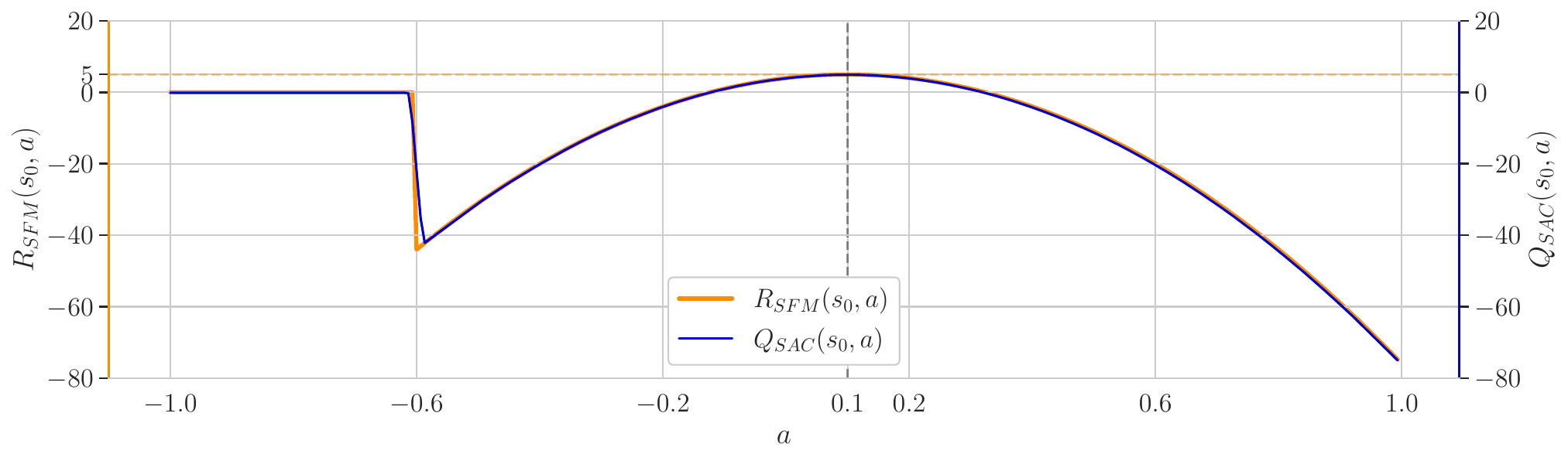}
    \caption{In orange: the reward function $R_{SFM}$ of the SFM environment. Since all transitions are terminal, $R_{SFM}$ coincides with the optimal Q-function. In blue: the critic ($Q_{SAC}$) obtained after a training of 20,000 steps with SAC \citep{haarnoja2018soft}.}
    \label{fig:SFM}
\end{figure}

\begin{figure}[htbp]
    \begin{subfigure}{0.5\linewidth}
        \includegraphics[width=\linewidth]{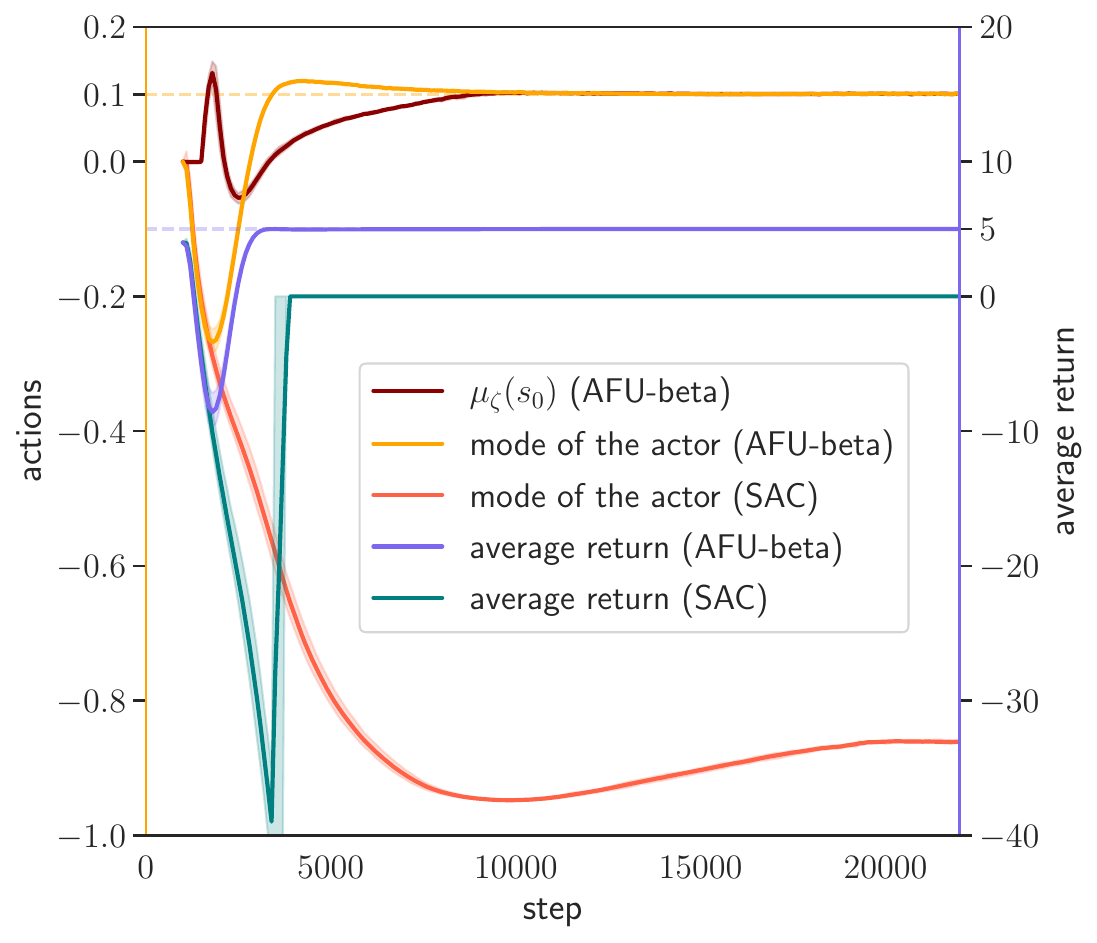}
        \caption{AFU-beta converges to the optimal solution, while SAC converges to a suboptimal solution. How to read the figure: the y-axis on the left (action) applies to the first three curves ($\mu_\zeta(s)$
        and the modes of the actor for AFU-beta and SAC), while the y-axis on the right (average return)
        applies for the two other curves (average returns for AFU-beta and SAC).}
        \label{fig:SAC_and_AFUbeta_on_SFM_sub1}
    \end{subfigure}
    \hspace*{\fill}
    \begin{subfigure}{0.5\linewidth}
        \includegraphics[width=\linewidth]{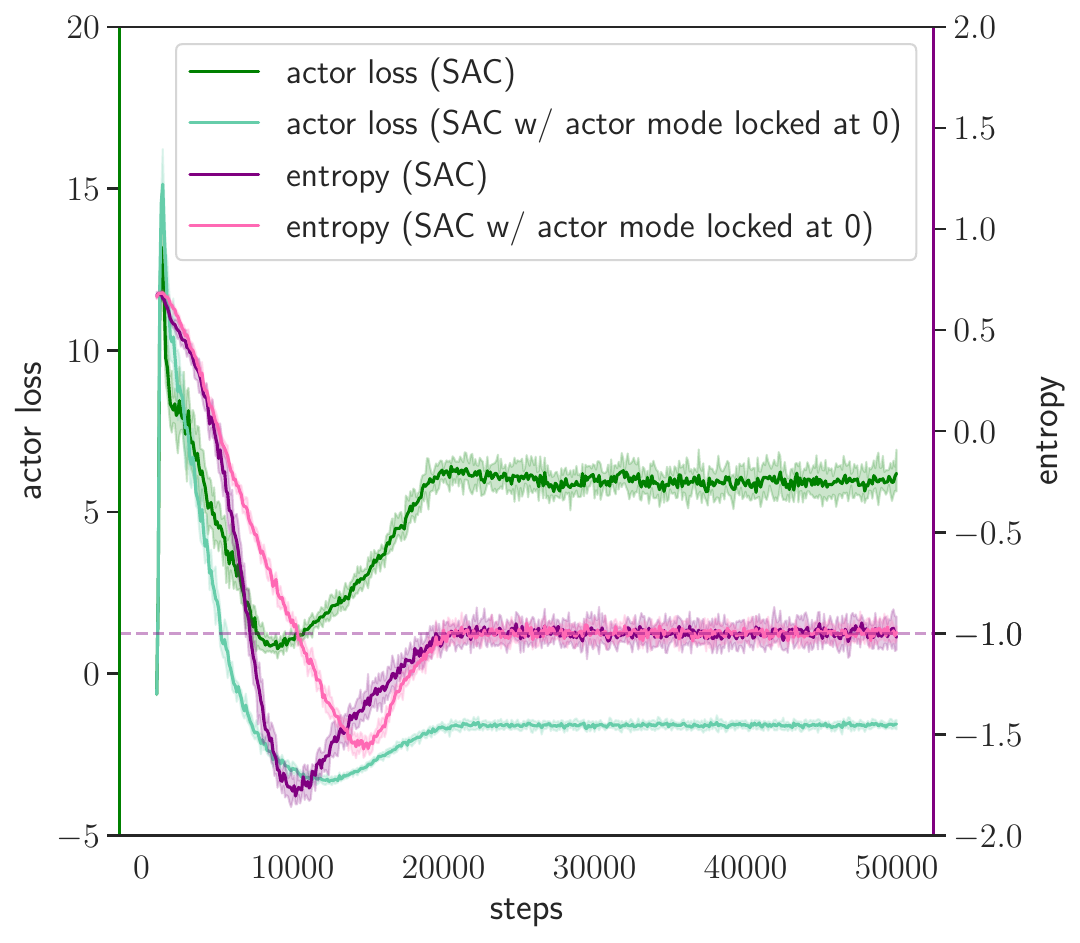}
        \caption{Evolution of the actor loss and entropy for SAC and SAC with the mode of its actor locked at $0$. The y-axis on the left (actor loss) applies to the first two curves (actor losses for both versions of SAC), and the y-axis on the right applies to the two other curves (entropies for both versions
        of SAC).}
        \label{fig:SAC_and_AFUbeta_on_SFM_sub2}
    \end{subfigure}
    \caption{Trainings of SAC and AFU-beta in the SFM environment. Plots show results averaged over 10 runs with different random seeds, and shaded areas range from the 25th to the 75th percentile.}
    \label{fig:SAC_and_AFUbeta_on_SFM}
\end{figure}

With a deterministic actor trained by stochastic gradient ascent over the Q-function landscape, DDPG, TD3 and similarly structured deterministic actor-critic algorithms can easily get stuck in local optima (see \cite{matheron2020understanding}). 
With a stochastic actor and updates based on the Kullback-Leibler (KL) divergence between output distributions and target distributions of the form $\exp\bigl(\frac{1}{\alpha}Q(s, \cdot)\bigr)/z(s)$, algorithms like SAC are less prone to deadlocks.
For instance, in areas where the gradient of the Q-function is close to zero, exploiting the KL loss results in an increase of the variance of the action distribution, which eventually helps find larger gradients and escape from the flat region.
Yet, the policy networks used in practice mostly output unimodal action distributions\footnote{This is starting to change, thanks to the influence of recent methods such as diffusion policies (see \citep{chi2023diffusion, DBLP:journals/corr/abs-2304-10573}), but such expressive and multimodal stochastic policies are still more cumbersome than unimodal policies.}, and with this restriction even the KL loss generates undesirable local optima. We illustrate this with a trivial environment which we call SFM (for ``SAC Failure Mode''). It consists of a single state $s_0$, and unidimensional actions in $[-1, 1]$. The reward of an action is given by the function:
\begin{align}
R_{SFM}(s_0, a) =\left\{
    \begin{array}{ll}
      5 - 100 (a - 0.1)^2, & \mbox{if $a \geq -0.6$},\\
      0, & \mbox{otherwise}.
    \end{array}
  \right.
\end{align}
All transitions are terminal, so all episodes stop after one step. The optimal policy selects $a=0.1$ and yields a return of $5$. We train SAC on SMF with the same hyperparameters as in our other experiments. We start by performing 1000 steps with random actions, which helps the critic $Q_{SAC}$ quickly converge toward the optimal $Q$-function, $Q^*$, which is simply equal to $R_{SFM}$. Figure~\ref{fig:SFM} shows $Q_{SAC}$ after 20,000 steps. Although $Q_{SAC}$ converges toward a very precise approximation of $Q^*$, the actor policy converges toward a suboptimal solution, as shown in Figure~\ref{fig:SAC_and_AFUbeta_on_SFM_sub1}. If we just modify SAC by locking the mode of the policy distribution at $0$, we can see in Figure~\ref{fig:SAC_and_AFUbeta_on_SFM_sub2} that the actor loss becomes much smaller, even after convergence of the actor entropy, which indicates that the policy of the default SAC algorithm gets stuck in a local optimum. There are two phases in the failure mode: at the beginning, when the entropy is relatively large, 
the asymmetry of $R_{SFM}$ makes the actor shift toward $-1$. As seen in Figure~\ref{fig:SFM}, $Q_{SAC}$ approximates the discontinuity in $R_{SFM}$ with a steep slope, and when the policy distribution becomes concentrated on the left of this slope, it acts as a barrier that traps the actor. Later in the training, when the entropy becomes smaller and converges to the target entropy ($-1$), it would be much preferable for the mode of the policy to converge back toward $0.1$, but the steep slope results in a deceptive gradient in the KL loss that prevents it from happening, and SAC remains stuck in the local optimum.

%% file: afubeta.tex
With the same actor loss as SAC, AFU-alpha fails similarly on SFM. We propose to improve the actor loss to make it less likely to get stuck in local optima.
The first idea is to train by regression an estimate of where the mode of the actor should be.
If the learning progresses well, $Q_\psi(s, a) > \min_{i \in \{1,2\}}\bigl(V_{\varphi_i}(s)\bigr)$ should only be possibly true in the vicinity of the argmax ($\text{argmax}_{a \in A}\bigl(Q(s,a)\bigr)$), so we use actions $a$ with a Q-value greater than $\min_{i \in \{1,2\}}\bigl(V_{\varphi_i}(s)\bigr)$ as targets. To find such actions we use both actions in the mini-batches and actions resampled with the actor on those mini-batches. Let us consider a mini-batch $B$ of transitions $(s, a, r, s')$, and actions $a_s$ resampled with the actor. We denote by $\mathcal{M}(B)$ the set of state-action pairs $(s, a_\bullet)$
such that $a_\bullet = a$ or \mbox{$a_\bullet = a_s$} and $Q_\psi(s, a_\bullet) > \min_{i \in \{1,2\}}\bigl(V_{\varphi_i}(s)\bigr)$.
We introduce a new deterministic function \mbox{approximator} \mbox{$\mu_\zeta: S \rightarrow A$} with parameters $\zeta$ and train it with the following loss:
\begin{align}
\label{eq:Lmu}
L_{\mu}(\zeta) = \underset{(s, a_\bullet) \in \mathcal{M}(B)}{\text{Mean}}\left[ \Bigl(\mu_\zeta(s) - a_\bullet\Bigr)^2 \right].
\end{align}
In our implementation, most of the parameters between $\zeta$ and $\theta$ are shared: we simply modify the output dimension of $\pi_\theta$ to make it also return $\mu_\zeta(s)$.
It does not change the approach in any way, but when computing the gradient of the loss~\eqref{eq:Lmu}, one must carefully ignore the influence of the parameters $\zeta$ on resampled actions $a_s$.

Let us reconsider the actor loss from Equation~\eqref{eq:Lpi}.
It balances two terms, the first one ($\alpha \log(\pi_\theta(a_s | s))$) that maximizes the entropy, and the second one ($-Q_\psi(s, a_s)$) that encourages $\pi_\theta$ to output
actions maximizing $Q_\psi(s, \cdot)$. In the gradient $\nabla_\theta L_{\pi}(\theta)$, which can be expressed by making explicit the relationship between sampled actions $a_s$ and the input noise (see \cite{haarnoja2018soft}), the second term results in small modifications of $\theta$ that attempt to change the actions $a_s$ in the direction of $\nabla_{a} Q_\psi(s, a)$, where $a$ is evaluated in $a_s$, and which we write by abuse of notation $\nabla_{a_s} Q_\psi(s, a_s)$. If $\nabla_{a_s} Q_\psi(s, a_s)$ points away from the global optimum, it can contribute to the creation of a local minimum in the actor loss. 
We want to edit $\nabla_{a_s} Q_\psi(s, a_s)$ in order to avoid deceptive gradients. To do so, we compute the dot product between $\nabla_{a_s} Q_\psi(s, a_s)$ and $\mu_\zeta(s) - a_s$, which is an estimate of a direction toward $\text{argmax}_{a \in A}\bigl(Q(s,a)\bigr)$. If the dot product is positive or zero, the gradient does not point away from $\mu_\zeta(s)$, so we can keep it unchanged. However, if $\nabla_{a_s} Q_\psi(s, a_s) \cdot (\mu_\zeta(s) - a_s) < 0$, then we project $\nabla_{a_s} Q_\psi(s, a_s)$ onto $(\mu_\zeta(s) - a_s)^\perp$ to anneal the dot product. We do it only if we estimate that $a_s$ is not already in the vicinity of the argmax, i.e. if $Q_\psi(s, a_s) < \min_{i \in \{1,2\}}\bigl(V_{\varphi_i}(s)\bigr)$. We introduce the following operator:
\begin{align}
\mathcal{G}^{s, a_s}(v) =\left\{
    \begin{array}{ll}
      \text{proj}_{(\mu_\zeta(s) - a_s)^\perp}\bigl(v\bigr), & \mbox{if $v \cdot (\mu_\zeta(s) - a_s) < 0$ and $Q_\psi(s, a_s) < \min_{i \in \{1,2\}}\bigl(V_{\varphi_i}(s)\bigr)$}.\\
      v, & \mbox{otherwise}.
    \end{array}
  \right.
\end{align}
When computing the gradient $\nabla_\theta L_{\pi}(\theta)$, we replace the terms $\nabla_{a_s} Q_\psi(s, a_s)$ (resulting from the chain rule) by $\mathcal{G}^{s, a_s}\bigl(\nabla_{a_s} Q_\psi(s, a_s)\bigr)$. It leads to a modified gradient which we denote by
$
\nabla^{\text{MODIF}}_\theta L_{\pi}(\theta).
$

\begin{figure}[htbp]
	\includegraphics[width=\linewidth]{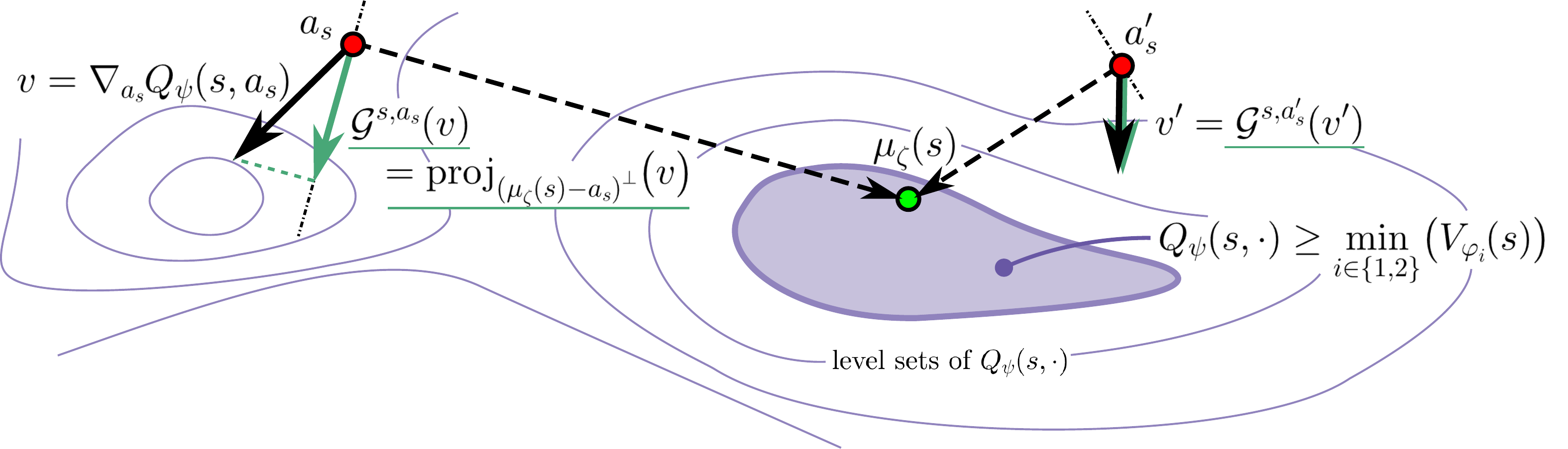}
    \caption{The gradient $v$ at $a_s$ (on the left) points away from $\mu_\zeta(s)$, which determines the direction toward the vicinity of the argmax of $Q_\psi(s, \cdot)$, so we modify $v$ to get $\mathcal{G}^{s, a_s}\bigl(v\bigr)$ by projecting it on the hyperplane orthogonal to $\mu_\zeta(s) - a_s$. The gradient $v'$ at $a'_s$ (on the right) points in the direction (half-space) of $\mu_\zeta(s)$, so we do not modify it, and $\mathcal{G}^{s, a'_s}\bigl(v'\bigr) = v'$.}
    \label{fig:illus2}
\end{figure}

This process is illustrated in Figure~\ref{fig:illus2}. It can be understood as an artificial modification of the landscape of $Q_\psi(s, \cdot)$ so that, outside the region defined by $Q_\psi(s, \cdot) \geq \min_{i \in \{1,2\}}\bigl(V_{\varphi_i}(s)\bigr)$, its gradient never points away from $\mu_\zeta(s)$. 
$\mu_\zeta(s)$ has the advantage of being trained by regression, and its training includes actions coming directly from the replay buffer, not only ones resampled by the actor. It means that, in a very off-policy setting, if a new peak of $Q_\psi(s, a)$ appears far from the actions currently likely to be sampled, the update of $\mu_\zeta(s)$ can occur first and then guide the update of $\pi_\theta$ by removing all deceptive gradients that would need to be crossed to reach the new peak.
More generally, the use of $\mu_\zeta$ prevents the actor from being trapped in local optima, as long as the training of the critic is doing well. Since training the critic is independent from the actor, we believe that our proposed approach goes one step further in the development of sound foundations for a purely off-policy reinforcement learning algorithm performing well in continuous action spaces.

We call AFU-beta the updated algorithm. It works like AFU-alpha, with the additional training of $\mu_\zeta$ and the replacement of $\nabla_\theta L_{\pi}(\theta)$ by $\nabla^{\text{MODIF}}_\theta L_{\pi}(\theta)$, as described in Algorithm~\ref{alg:afualpha}.
Figure~\ref{fig:SAC_and_AFUbeta_on_SFM_sub1} shows that, in the SFM environment, unlike SAC, AFU-beta quickly converges to the optimal solution.

We evaluate AFU-beta on the MuJoCo benchmark in the same way as AFU-alpha and show results in Figure~\ref{fig:Benchmark} (Appendix~\ref{app:beta}).
Again, AFU-beta is competitive with SAC and TD3. The differences between AFU-beta and AFU-alpha are not very significant on the MuJoCo benchmark, possibly because issues with local optima are rarely encountered in these environments. We leave for future work the search for meaningful and complex environments in which AFU-beta has a notable advantage over AFU-alpha.

%% file: ccl.tex
We presented AFU, an off-policy RL algorithm with critic updates independent from the actor. 
At its core is a novel way to solve the continuous action Q-function maximization (max-Q) problem using regression and conditional gradient scaling, which we believe could have applications outside the field of reinforcement learning. 
%We plan to work on a mathematical proof of the convergence of this new method and analyze whether connections with it can be established in the vast landscape of convex and non-convex optimization, for instance with penalty and multiplier methods \citep{bertsekas1976multiplier} or Thikonov regularization \citep{tikhonov1963solution}.

The first version of AFU (AFU-alpha) has a stochastic actor trained as in SAC \citep{haarnoja2018soft}. We provide a simple example of failure mode for SAC, and show how the value function trained in AFU can help improve the actor loss and make it less prone to local optima, resulting in a second version of AFU (AFU-beta) which does not exhibit the same failure mode as SAC.

Our experimental results on a classical benchmark show that both versions of AFU are competitive with SAC and TD3, two state-of-the-art off-policy model-free RL algorithms. As far as we know, AFU is the first off-policy RL algorithm that is competitive in sample-efficiency with the state-of-the-art and truly departs from the actor-critic perspective. We believe that it could open up new avenues for off-policy RL algorithms applied to continuous control problems.